\pdfoutput=1
\documentclass[10pt,twocolumn,letterpaper]{article}

\usepackage[letterpaper,textwidth=6.75in,textheight=9.25in, columnsep=0.25in,centering]{geometry}
\usepackage[T1]{fontenc}
\usepackage[utf8]{inputenc}
\usepackage{times}          
\usepackage{authblk}

\usepackage{amsmath}
\usepackage{amssymb}
\usepackage{booktabs}
\usepackage{tabularx}
\usepackage{array}
\usepackage{graphicx}
\usepackage{float}
\usepackage{mathtools}
\usepackage{parskip}

\DeclareMathOperator*{\argmin}{arg\,min}

\usepackage[round]{natbib}
\usepackage[colorlinks=true,linkcolor=blue,citecolor=blue,urlcolor=blue]{hyperref}

\title{Evaluating Bayesian deep learning for robust biomedical time-series
classification under distribution shift: A sleep scoring case study}

\author[1,2]{Mads Birch Sørensen}
\author[2]{Ulrike Richter}
\author[2]{Francois Gastambide}
\author[1]{Michael Riis Andersen}

\affil[1]{Technical University of Denmark, Denmark}
\affil[2]{H.\ Lundbeck A/S, Denmark}

\date{}

\begin{document}
\maketitle

\begin{abstract}
In biomedical time-series classification, machine learning models often suffer from deteriorating accuracy and calibration due to distribution shift. We investigate to what degree Bayesian-inspired deep learning improves robustness through Bayesian model averaging (BMA). Using rodent sleep scoring as a case study, we evaluate BMA under distribution shifts in genotype, laboratory, and species. We systematically compare nine approximate inference methods spanning single- and multi-basin approaches across two neural network backbones on in-distribution (ID) data and five out-of-distribution (OOD) datasets. In our results, multi-basin methods perform best, outperforming baselines and single-basin methods. Compared with point estimates, BMA generally improves OOD accuracy and calibration across both architectures, while yielding only modest ID gains. It also often improves selective prediction and OOD and artifact detection, but, interestingly,  not inter-rater disagreement detection.
\end{abstract}

\section{INTRODUCTION}

Machine learning has demonstrated high predictive accuracy across a wide range of biomedical time-series classification tasks, including cardiac rhythm classification from electrocardiograms \citep{Hannun.etal-2019}, seizure detection and seizure-type classification from electroencephalographic recordings \citep{Tang.etal-2022}, and sleep staging from electrophysiological recordings \citep{Miladinovic.etal-2019,Perslev.etal-2021}. However, in practical applications, models are often applied to data from different populations, recording sites, sensor types and placements, or annotation practices, resulting in distribution shift relative to the training data \citep{Ovadia.etal-2019,Dockes.etal-2021,Wagh.etal-2022}. Moreover, several of these factors may change simultaneously, producing structured shifts that may not be well represented by the synthetic corruptions commonly used to study distribution shift \citep{Koh.etal-2021}. Generalization beyond the training distribution is therefore critical for reliable deployment.

As both predictive accuracy and model confidence often degrade under distribution shift \citep{Ovadia.etal-2019}, reliable deployment necessitates well-calibrated probabilities to flag potential model errors. A well-calibrated classifier assigns confidence levels that match the observed frequency of correct predictions \citep{Guo.etal-2017}. Such probabilities can support risk-aware use of machine-learning models, including selective prediction, in which uncertain cases are deferred for expert review \citep{Amodei.etal-2016,Geifman.El-Yaniv-2017}.

Bayesian neural networks represent model uncertainty by inferring a posterior distribution over the model parameters or the function space directly and form predictions by marginalizing over the corresponding posterior \citep{Blundell.etal-2015,Rudner.etal-2022}. Because exact Bayesian inference is generally intractable for neural networks, practical Bayesian model averaging (BMA) relies on scalable approximations. Many of these approaches characterize uncertainty within a single posterior basin, whereas deep ensembles (DEs) and related multi-basin methods average predictions across independent solutions \citep{Lakshminarayanan.etal-2017,Wilson.Izmailov-2020}.

However, whether approximate BMA systematically improves reliability under distribution shift remains unsettled. Large-scale comparisons have found that model averaging, particularly DEs, can provide strong predictive performance and uncertainty estimates under increasing distribution shift \citep{Ovadia.etal-2019}. However, a more accurate approximation of the Bayesian posterior does not necessarily translate into greater robustness under covariate shift \citep{Izmailov.etal-2021}, and adding local posterior structure to DEs can improve out-of-distribution (OOD) performance while reducing in-distribution (ID) performance \citep{Jordahn.etal-2025}. At the same time, most prior work on Bayesian deep learning relies on standard benchmarks and vision applications \citep{Papamarkou.etal-2024}. Whether these findings extend to naturally occurring distribution shifts in biomedical time-series classification is therefore an open question.

We address this question using automated rodent sleep scoring from electrophysiological recordings as a case study. Several strong sleep-scoring models already exist \citep{Miladinovic.etal-2019,Liu.etal-2022,Brodersen.etal-2024}, yet the task combines several challenges relevant to reliable deployment. Recordings vary across laboratories, acquisition systems, genotypes, experimental interventions, and species, with several factors varying simultaneously across cohorts \citep{Miladinovic.etal-2019}. Moreover, rodent sleep scoring lacks universally accepted scoring conventions, and signal artifacts can complicate annotation \citep{Rayan.etal-2024a}. These characteristics make rodent sleep scoring a useful test case for evaluating approximate BMA under structured, naturally occurring distribution shifts and for assessing whether predictive uncertainty identifies OOD recordings, artifacts, and disagreement between independent scorers. Although prior work has examined generalization across several of these shifts, to our knowledge no sleep-staging study has systematically compared a broad range of single- and multi-basin approximate BMA methods under natural distribution shift.

Our contributions are as follows:
\begin{itemize}
    \item We conduct a systematic comparison of nine scalable posterior approximation and model-averaging methods across two complementary neural network architectures. The methods are the last-layer Laplace approximation (LLLA) \citep{Kristiadi.etal-2020}; stochastic weight averaging (SWA) and stochastic weight averaging-Gaussian (SWAG) \citep{Izmailov.etal-2018,Maddox.etal-2019}; variational gradient descent (VGD) and mean-field variational inference (MFVI) \citep{Ghosh.etal-2025,Blundell.etal-2015}; stochastic gradient Langevin dynamics (SGLD) \citep{Welling.Teh-2011}; DEs, MultiSWA, and MultiSWAG \citep{Lakshminarayanan.etal-2017,Wilson.Izmailov-2020}.
    
    \item We evaluate classification accuracy and calibration for a large in-the-wild dataset and five naturally shifted OOD cohorts spanning differences in genotype, laboratory and acquisition configuration, signal montage, and species.

    \item Beyond aggregate predictive performance and calibration, we assess whether predictive uncertainty supports OOD, artifact, and inter-scorer disagreement detection, as well as selective prediction.
\end{itemize}

\section{RELATED WORK}

\paragraph{Automatic sleep staging and generalization}
Sleep staging has been approached using several model architectures. In human sleep staging, prominent examples include convolutional neural networks (CNNs) \citep{Perslev.etal-2021}, convolutional-recurrent models \citep{Supratak.etal-2017,Olesen.etal-2021}, transformer-based models \citep{Phan.etal-2022,Guo.etal-2024}, and, more recently, foundation models with sleep staging as a downstream task \citep{Thapa.etal-2026}. Rodent sleep-scoring models span similar architectures, including convolutional-recurrent models \citep{Yamabe.etal-2019,Smith.etal-2025}, temporal convolutional networks \citep{Liu.etal-2022}, and classifiers that combine discriminative models with a hidden Markov model (HMM) to capture sleep-stage transitions \citep{Miladinovic.etal-2019,Brodersen.etal-2024}. Across these architectures, modern sleep-staging models have achieved high classification performance on their respective evaluation datasets.

Several studies have investigated robustness across cohorts, populations, recording sites, acquisition systems, and signal montages in human sleep staging \citep{Perslev.etal-2021,Guillot.etal-2020,Guillot.Thorey-2021,Guo.etal-2024}, as well as across experimental conditions, recording configurations, laboratories, genotypes, and species in rodent sleep scoring \citep{Barger.etal-2019,Miladinovic.etal-2019,Liu.etal-2022,Brodersen.etal-2024}. Despite these efforts, generalization beyond the source domain remains a central challenge and an important barrier to widespread clinical adoption \citep{Rossi.etal-2025}.

\paragraph{Uncertainty quantification in sleep staging}
Sleep staging can be ambiguous, as reflected by disagreement between expert scorers \citep{Guillot.etal-2020,Bakker.etal-2023,Fiorillo.etal-2023a}. Probabilistic predictions provide a natural way to represent this ambiguity by assigning probabilities across sleep stages rather than a single hard label \citep{Stephansen.etal-2018}. This motivates a distinction between \textit{aleatoric} uncertainty arising from ambiguity in the data and \textit{epistemic} uncertainty arising from limited model knowledge \citep{Kendall.Gal-2017, vanGorp.etal-2022}. Uncertainty estimates have also been used to flag ambiguous predictions for human review \citep{Brodersen.etal-2024,Heremans.etal-2024}.

A smaller body of work has used approximate BMA to quantify epistemic uncertainty, generally focusing on one or a small number of posterior approximation methods. Recent studies have predominantly used MC dropout \citep{Fiorillo.etal-2021,Vainikka.etal-2026,Yu.etal-2026}, while \citet{Yu.etal-2026} additionally use variational inference and model ensembling. In related work, \citet{Rossi.etal-2025} use disagreement within a model ensemble to identify ambiguous signal windows.

\paragraph{Bayesian model averaging under distribution shift}
Prior work shows that the benefits of approximate BMA under distribution shift depend on both method and evaluation setting. \citet{Ovadia.etal-2019} found that predictive accuracy and uncertainty quality deteriorate with increasing shift and that DEs are generally among the most robust approaches. SWAG has shown strong calibration and OOD detection \citep{Maddox.etal-2019}, while MultiSWAG can outperform DEs and MultiSWA under severe image corruptions \citep{Wilson.Izmailov-2020}. Evaluations on realistic shifts similarly find that multi-basin approximations often improve generalization and calibration, although the gains depend on the task and architecture \citep{Seligmann.etal-2023}. Moreover, adding local posterior structure to DEs can improve OOD metrics at the cost of ID performance \citep{Jordahn.etal-2025}, and more accurate posterior approximation does not necessarily improve robustness under covariate shift \citep{Izmailov.etal-2021}. 

However, much of the empirical Bayesian deep-learning literature focuses on standard benchmarks \citep{Papamarkou.etal-2024}, while sleep-staging studies have not compared a wide range of approximate BMA methods under distribution shift. Together, these gaps motivate evaluating single- and multi-basin approximate BMA methods in rodent sleep scoring, where several naturally occurring shifts provide a challenging test of robustness.

\section{BACKGROUND \& METHOD}
Sleep scoring aims to identify sleep stages over time from electrophysiological recordings. We formulate this as a multi-class sequence-to-sequence classification problem. Each multi-channel recording is divided into windows of $T$ samples and is assigned a discrete sleep score label $y \in \mathcal{Y}$. We consider a dataset with $N_{\mathrm{seq}}$ sequences
$
\mathcal{D}
=
\{(\mathbf{X}_n,\mathbf{y}_n)\}_{n=1}^{N_{\mathrm{seq}}},$
where $\mathbf{X}_n\in\mathbb{R}^{C\times WT}$ contains $W$ consecutive windows from $C$ electrophysiological channels, concatenated along the
temporal dimension, and $\mathbf{y}_n=(y_{n,1},\ldots,y_{n,W})\in\mathcal{Y}^{W}$ contains the corresponding sleep-stage labels.

\paragraph{Model and likelihood}
For each window $w$ in an input sequence $\mathbf{X}_n$, we parameterize the class probabilities using a neural network $f_{\boldsymbol{\theta}}$ with parameters $\boldsymbol{\theta}$:
$$
p(y_{n,w}\mid\mathbf{X}_n,\boldsymbol{\theta})
=
\operatorname{Cat}\!\left(
y_{n,w}\mid f_{\boldsymbol{\theta}}(\mathbf{X}_n)_w
\right),
$$
where $f_{\boldsymbol{\theta}}(\mathbf{X}_n)_w$ denotes the probability vector over sleep-stage classes for window $w$. Assuming conditional independence of the output labels given the input and model parameters, the likelihood factorizes as
\begin{align}
p(\mathcal{D}\mid\boldsymbol{\theta})
=
\prod_{n=1}^{N_{\mathrm{seq}}}
\prod_{w=1}^{W}
p(y_{n,w}\mid\mathbf{X}_n,\boldsymbol{\theta}).
\label{eq:likelihood}
\end{align}
Although the likelihood factorizes over windows, each factor is conditioned on the complete input sequence, allowing the model to use temporal context from neighboring windows.

\paragraph{The posterior predictive distributions}
Given a prior distribution $p(\boldsymbol{\theta})$, Bayes' rule defines the posterior distribution
\begin{equation}
p(\boldsymbol{\theta}\mid\mathcal{D})
= \frac{p(\mathcal{D}\mid\boldsymbol{\theta})
p(\boldsymbol{\theta})}{p(\mathcal{D})}
\propto e^{-U(\boldsymbol{\theta})},
\label{eq:bayes}
\end{equation}
where $U(\boldsymbol{\theta}) = -\log p(\boldsymbol{\theta}) - \log p(\mathcal{D}\mid\boldsymbol{\theta})$ is the potential energy. 
While the maximum a posteriori (MAP) estimate is tractable using stochastic optimization,
$
\hat{\boldsymbol{\theta}}_{\mathrm{MAP}} = \argmin_{\boldsymbol{\theta}} U(\boldsymbol{\theta})
$, exact posterior inference is generally intractable for neural networks, so we resort to approximate inference, where $p(\boldsymbol{\theta}\mid\mathcal{D})\approx q_m(\boldsymbol{\theta})$ denotes the approximate distribution for method $m$. 

For a new input sequence $\mathbf{X}^*$, the posterior predictive distribution is obtained through approximate BMA
\begin{equation}
\begin{aligned}
p(\mathbf{y}^*\mid\mathbf{X}^*,\mathcal{D})
&\approx
\int
p(\mathbf{y}^*\mid\mathbf{X}^*,\boldsymbol{\theta})
q_m(\boldsymbol{\theta})
\,\mathrm{d}\boldsymbol{\theta}
\\
&\approx
\frac{1}{S}
\sum_{s=1}^{S}
p(\mathbf{y}^*\mid\mathbf{X}^*,\boldsymbol{\theta}^{(s)}),
\end{aligned}
\label{eq:predictive}
\end{equation}
where $\boldsymbol{\theta}^{(s)}\sim q_m(\boldsymbol{\theta})$ are samples from an approximate posterior distribution and $S$ is the number of samples.

\subsection{Posterior approximation methods}
\label{sec:inference}

We compare the MAP reference, i.e., the point estimate $\hat{\boldsymbol{\theta}}_{\text{MAP}},$ with nine scalable approximate inference methods. Table~\ref{tab:posteriors} summarizes the method-specific posterior distributions $q_m(\boldsymbol{\theta})$ used in Eq.~\eqref{eq:predictive}. Except for VGD and DEs, we treat all methods as post-hoc methods initialized at the MAP solution $\hat{\boldsymbol{\theta}}_{\mathrm{MAP}}$.

\paragraph{LLLA}
The Laplace approximation approximates the posterior distribution using a Gaussian centered at the MAP estimate, with covariance determined by the local curvature of $U(\boldsymbol{\theta})$ \citep{Daxberger.etal-2021}. Computing and storing full-network curvature is infeasible for large neural networks. We therefore use LLLA, which fixes the non-last-layer parameters $\boldsymbol{\theta}^{-\mathrm{LL}}$ at their MAP values and applies the Gaussian approximation only to the last-layer parameters $\boldsymbol{\theta}^{\mathrm{LL}}$ \citep{Kristiadi.etal-2020}. The last-layer curvature matrix is
$$
\boldsymbol{\mathcal{H}}^{\mathrm{LL}}
=
-\left.
\nabla_{\boldsymbol{\theta}^{\mathrm{LL}}}^{2}
\log p\!\left(
    \mathcal{D},
    \boldsymbol{\theta}_{\mathrm{MAP}}^{-\mathrm{LL}},
    \boldsymbol{\theta}^{\mathrm{LL}}
\right)
\right|_{\boldsymbol{\theta}^{\mathrm{LL}}
=
\boldsymbol{\theta}_{\mathrm{MAP}}^{\mathrm{LL}}},
$$
or a positive-semidefinite approximation thereof, and the corresponding covariance is
$
\boldsymbol{\Sigma}^{\mathrm{LL}}
=
\left(
\boldsymbol{\mathcal{H}}^{\mathrm{LL}}
\right)^{-1}.
$

\paragraph{SWA and SWAG}
\citet{Mandt.etal-2017} provide a Bayesian interpretation of constant-learning-rate stochastic gradient descent (SGD), in which iterates from the stationary regime are viewed as samples from an approximately Gaussian distribution over model parameters. Let 
$
\boldsymbol{\theta}_{\mathrm{MAP}},\ldots,\boldsymbol{\theta}_G
$ 
denote $G$ parameter vectors collected along such an SGD trajectory starting at the MAP solution. SWA estimates the mean of this distribution by averaging the iterates \citep{Izmailov.etal-2018}
$
\boldsymbol{\theta}_{\mathrm{SWA}}
=
\frac{1}{G}
\sum_{g=1}^{G}
\boldsymbol{\theta}_g.
$
SWAG extends this by additionally estimating the covariance of the SGD iterates, yielding a Gaussian approximation centered at $\boldsymbol{\theta}_{\mathrm{SWA}}$ \citep{Maddox.etal-2019}. The covariance combines a diagonal component estimated from all $G$ iterates,
$
\boldsymbol{\Sigma}_{\mathrm{diag}}
=
\operatorname{diag}\!\left(
\overline{\boldsymbol{\theta}^{2}}
-
\boldsymbol{\theta}_{\mathrm{SWA}}^{2}
\right),
$
where
$
\overline{\boldsymbol{\theta}^{2}}
=
\frac{1}{G}
\sum_{g=1}^{G}
\boldsymbol{\theta}_g^{2},
$
and the squaring operations are element-wise and a low-rank component
$$
\boldsymbol{\Sigma}_{\mathrm{low\text{-}rank}}
=
\frac{1}{R-1}
\sum_{g=G-R+1}^{G}
\left(
\boldsymbol{\theta}_g-\bar{\boldsymbol{\theta}}_R
\right)
\left(
\boldsymbol{\theta}_g-\bar{\boldsymbol{\theta}}_R
\right)^{\top},
$$
where $\bar{\boldsymbol{\theta}}_R$ is the mean of the final $R$ iterates. The resulting SWAG covariance is then 
$
\boldsymbol{\Sigma}_{\mathrm{SWAG}}
=
\frac{1}{2}
\left(
\boldsymbol{\Sigma}_{\mathrm{diag}}
+
\boldsymbol{\Sigma}_{\mathrm{low\text{-}rank}}
\right).
$

\paragraph{VGD and MFVI}
Variational inference approximates $p(\boldsymbol{\theta}\mid\mathcal{D})$ with a tractable distribution $q_{\mathrm{VI}}(\boldsymbol{\theta})$ by minimizing the negative evidence lower bound,
\begin{equation*}
\mathcal{L}_{\mathrm{VI}}
=
-
\mathbb{E}_{q_{\mathrm{VI}}(\boldsymbol{\theta})}
\left[
\log p(\mathcal{D}\mid\boldsymbol{\theta})
\right]
+
\operatorname{KL}\!\left[
q_{\mathrm{VI}}(\boldsymbol{\theta})
\,\|\,p(\boldsymbol{\theta})
\right].
\label{eq:vi-objective}
\end{equation*}
MFVI assumes a fully factorized Gaussian distribution and jointly learns the mean and marginal variance of each parameter \citep{Blundell.etal-2015}. A related method is VGD, which is designed to find flatter minima for better generalization  \citep{Ghosh.etal-2025}. It uses an isotropic Gaussian with fixed scalar variance $\sigma^2 >0$ and optimizes only its mean $\boldsymbol{\mu}$. For fixed $\sigma^2$, its variational objective is equivalent, up to a constant, to minimizing the MAP objective averaged over Gaussian parameter perturbations:
\begin{equation*}
\mathcal{L}_{\mathrm{VGD}}(\boldsymbol{\mu})
=
\mathbb{E}_{\boldsymbol{\epsilon}
\sim
\mathcal{N}(\mathbf{0},\sigma^2\mathbf{I})}
\left[
U
\left(
\boldsymbol{\mu}+\boldsymbol{\epsilon}
\right)
\right],
\label{eq:vgd-objective}
\end{equation*}
where the expectation is approximated using sampling.

\paragraph{SGLD}
SGLD approximates posterior sampling by adding Gaussian noise to stochastic-gradient updates of the posterior energy \citep{Welling.Teh-2011}:
\begin{equation}
\boldsymbol{\theta}_{k+1}
=
\boldsymbol{\theta}_k
-
\eta
\nabla_{\boldsymbol{\theta}}
\widehat{U}_{\mathcal{B}}
\left(
\boldsymbol{\theta}_k
\right)
+
\sqrt{2\eta}\,
\boldsymbol{\epsilon}_k,
\label{eq:sgld}
\end{equation}
where $\boldsymbol{\epsilon}_k \sim \mathcal{N}(\mathbf{0},\mathbf{I})$ and $\eta>0$ is the step size and $\nabla_{\boldsymbol{\theta}}\widehat{U}_{\mathcal{B}}$ is a mini-batch estimate of $\nabla_{\boldsymbol{\theta}}U(\boldsymbol{\theta})$. After warm up, retained parameter iterates are treated as approximate posterior samples.

\paragraph{Multi-basin methods}
DEs average predictions from $M$ independently initialized models and can be viewed as an empirical multi-basin approximation to BMA \citep{Lakshminarayanan.etal-2017,Wilson.Izmailov-2020}. MultiSWA similarly combines independently trained SWA solutions, while MultiSWAG combines local SWAG approximations across multiple basins \citep{Wilson.Izmailov-2020}.

\begin{table}[t]
\centering
\caption{
Method-specific parameter distributions $q_m(\boldsymbol{\theta})$ used in Eq.~\eqref{eq:predictive}. $\theta^{LL}$ and $\theta^{-\mathrm{LL}}$ denote the last-layer and non-last-layer parameters, respectively. Point estimates are represented as Dirac measures $\delta$.}
\label{tab:posteriors}
\small
\setlength{\tabcolsep}{0pt}
\begin{tabularx}{\columnwidth}{
    @{}
    >{\raggedright\arraybackslash}p{0.33\columnwidth}
    >{\centering\arraybackslash}X
    @{}
}
\toprule
Method & $q_m(\boldsymbol{\theta})$ \\
\midrule
MAP & $\displaystyle\delta\!\left(\boldsymbol{\theta}-\boldsymbol{\theta}_{\mathrm{MAP}}\right)$ \\
\addlinespace[3pt]
LLLA & $\displaystyle\delta\!\left(\boldsymbol{\theta}^{-\mathrm{LL}}-\boldsymbol{\theta}_{\mathrm{MAP}}^{-\mathrm{LL}}\right)\mathcal{N}\!\left(\boldsymbol{\theta}^{\mathrm{LL}}\mid\boldsymbol{\theta}_{\mathrm{MAP}}^{\mathrm{LL}},\boldsymbol{\Sigma}^{\mathrm{LL}}\right)$ \\
\addlinespace[3pt]
SWA & $\displaystyle\delta\!\left(\boldsymbol{\theta}-\boldsymbol{\theta}_{\mathrm{SWA}}\right)$ \\
\addlinespace[3pt]
SWAG & $\displaystyle\mathcal{N}\!\left(\boldsymbol{\theta}\mid \boldsymbol{\theta}_{\mathrm{SWA}},\boldsymbol{\Sigma}_{\mathrm{SWAG}}\right)$ \\
\addlinespace[3pt]
VGD & $\displaystyle\mathcal{N}\!\left(\boldsymbol{\theta}\mid\boldsymbol{\mu},\sigma^{2}\mathbf{I}\right)$ \\
\addlinespace[3pt]
MFVI & $\displaystyle\mathcal{N}\!\left(\boldsymbol{\theta}\mid\boldsymbol{\mu},\operatorname{diag}\!\left(\boldsymbol{\sigma}^{2}\right)\right)$ \\
\addlinespace[3pt]
SGLD &$\displaystyle \frac{1}{S} \sum_{s=1}^{S} \delta\!\left( \boldsymbol{\theta} - \boldsymbol{\theta}^{(s)} \right)$ \\
\addlinespace[3pt]
DE & $\displaystyle\frac{1}{M}\sum_{j=1}^{M}\delta\!\left(\boldsymbol{\theta}-\boldsymbol{\theta}_{\mathrm{MAP}}^{(j)}\right)$ \\
\addlinespace[3pt]
MultiSWA & $\displaystyle\frac{1}{M}\sum_{j=1}^{M}\delta\!\left(\boldsymbol{\theta}-\boldsymbol{\theta}_{\mathrm{SWA}}^{(j)}\right)$ \\
\addlinespace[3pt]
MultiSWAG & $\displaystyle\frac{1}{M}\sum_{j=1}^{M}\mathcal{N}\!\left(\boldsymbol{\theta}\mid\boldsymbol{\theta}_{\mathrm{SWA}}^{(j)},\boldsymbol{\Sigma}_{\mathrm{SWAG}}^{(j)}\right)$ \\
\bottomrule
\end{tabularx}
\end{table}

\section{EXPERIMENTS \& RESULTS}
We first describe the experimental setup, then compare the MAP backbones with existing sleep-scoring methods and evaluate posterior approximations for both ID and OOD data. Finally, we assess predictive uncertainty for OOD, artifact, and scorer-disagreement detection and for selective prediction.

\subsection{Experimental setup}
\label{sec:exp-setup}

\paragraph{Data and models}
All models were trained on a large source-domain dataset and evaluated on an ID test set and five smaller OOD cohorts exhibiting natural shifts in laboratory, signal types, genotype, and species. The source-domain dataset comprises recordings from wild-type mice (WT), whereas the OOD datasets comprise a cohort of narcolepsy type 1 (NT1) mice and four cohorts from the \textsc{spindle} dataset of \citet{Miladinovic.etal-2019}, collected across three external laboratories. The WT dataset was split at the animal level into training, validation, and test sets using a 70/15/15\% split to avoid leakage between sets. Dataset details are summarized in Table~\ref{tab:datasets}, with additional information provided in Appendix Table~\ref{tab:datasets-full}.

Each recording provided three electrophysiological input channels, with the specific channel types differing across cohorts (Appendix Table~\ref{tab:datasets-full}). All recordings were resampled to $f_s=128$~Hz and divided into non-overlapping windows of duration $\tau=4s$, corresponding to $T=\tau f_s=512$ samples per window. The WT and NT1 datasets were annotated by a single scorer with labels $\mathcal{Y}=\{\text{wake},\text{NREM},\text{REM}\}$, whereas the \textsc{spindle} datasets were annotated independently by two scorers, who also flagged stage-specific artifacts. For the artifact- and inter-scorer-disagreement analyses, the original artifact indicators and both sets of annotations were retained. For all other experiments, artifact labels were mapped to their corresponding sleep stage, metrics were computed separately against each scorer, and averaged. Appendix~\ref{app:exp-setup} provides all details on the signals, annotations, and preprocessing.

\begin{table}[t]
\centering
\caption{
Datasets and evaluation domains.
$N_{\mathrm{rec}}$ denotes the number of 24\,h recordings.
WT is the source domain; NT1 and the \textsc{spindle} cohorts are held-out
OOD cohorts. The final column reports the primary shift relative to WT;
cohorts may also differ in acquisition system, signal montage, and
annotation protocol.
}
\label{tab:datasets}
\footnotesize
\setlength{\tabcolsep}{3pt}
\begin{tabularx}{\columnwidth}{
    @{}
    l
    l
    l
    r
    >{\raggedright\arraybackslash}X
    @{}
}
\toprule
Cohort & Use & Species & $N_{\mathrm{rec}}$ & Primary shift \\
\midrule
WT & Train    & Mouse & 169 & --- \\
WT & Val.     & Mouse & 36  & --- \\
WT & ID test  & Mouse & 36  & None \\
NT1 & OOD test & Mouse & 28 & Genotype \\
\addlinespace[4pt]
\textsc{spindle}-A & OOD test & Mouse & 4 & Lab \\
\textsc{spindle}-D & OOD test & Mouse & 6 & Lab \\
\textsc{spindle}-B & OOD test & Mouse & 4 & Lab $+$ genotype \\
\textsc{spindle}-C & OOD test & Rat   & 8 & Lab $+$ species \\
\bottomrule
\end{tabularx}
\end{table}

We evaluated four sleep-scoring baselines. Two were established rodent sleep-scoring methods: \textit{Spindle}, which combines a CNN with an HMM \citep{Miladinovic.etal-2019}, and \textit{Somnotate}, which combines linear discriminant analysis (LDA) with an HMM \citep{Brodersen.etal-2024}. We followed the original methods as closely as possible, see Appendices~\ref{app:spindle} and~\ref{app:somnotate} for details. The other two baselines were the MAP versions of the neural network backbones used to compare inference methods: U-Sleep \citep{Perslev.etal-2021}, adapted from human sleep scoring, and a CNN-GRU sequence model combining convolutional feature extraction with recurrent temporal modeling \citep{Yamabe.etal-2019,Olesen.etal-2021,Smith.etal-2025}. Appendix~\ref{app:determin-backbones} provides further details, and Table~\ref{tab:final-map-models} summarizes the final architectures.

\begin{table}[t]
\centering
\caption{\textbf{In-distribution accuracy and calibration} on the WT test set. Accuracy-type metrics in \%, LPD in nats; mean $\pm$ 95\% CI over training seeds ($t$-based); for DE, MultiSWA and MultiSWAG, mean $\pm$ half the 2.5--97.5\% interval over 100 bootstrap draws of ten members; \textbf{bold} = best in the column over all models, \underline{underline} = best within its section (U-Sleep, CNN-GRU, Other baselines).}
\label{tab:id-results}
\scriptsize
\setlength{\tabcolsep}{4pt}
\begin{tabular}{lcccc}
\toprule
Model & MF1\,$\uparrow$ & $\kappa$\,$\uparrow$ & LPD\,$\uparrow$ & Brier\,$\downarrow$ \\
\midrule
\multicolumn{5}{@{}l}{\itshape U-Sleep} \\
MAP & $89.7\,{\scriptstyle\pm1.5}$ & $88.6\,{\scriptstyle\pm3.1}$ & ${\scriptstyle-}.186\,{\scriptstyle\pm.048}$ & $9.4\,{\scriptstyle\pm2.6}$ \\
Temp. & $89.7\,{\scriptstyle\pm1.5}$ & $88.6\,{\scriptstyle\pm3.1}$ & ${\scriptstyle-}.185\,{\scriptstyle\pm.046}$ & $9.4\,{\scriptstyle\pm2.5}$ \\
SWA & $91.5\,{\scriptstyle\pm0.3}$ & $91.4\,{\scriptstyle\pm0.7}$ & ${\scriptstyle-}.141\,{\scriptstyle\pm.009}$ & $7.1\,{\scriptstyle\pm0.5}$ \\
SWAG & $91.7\,{\scriptstyle\pm0.2}$ & $91.6\,{\scriptstyle\pm0.5}$ & ${\scriptstyle-}.138\,{\scriptstyle\pm.005}$ & $6.9\,{\scriptstyle\pm0.3}$ \\
Laplace & $89.7\,{\scriptstyle\pm1.5}$ & $88.6\,{\scriptstyle\pm3.1}$ & ${\scriptstyle-}.186\,{\scriptstyle\pm.048}$ & $9.4\,{\scriptstyle\pm2.6}$ \\
VGD & $90.6\,{\scriptstyle\pm1.4}$ & $89.4\,{\scriptstyle\pm3.4}$ & ${\scriptstyle-}.181\,{\scriptstyle\pm.044}$ & $9.1\,{\scriptstyle\pm2.6}$ \\
MFVI & $90.5\,{\scriptstyle\pm0.6}$ & $89.7\,{\scriptstyle\pm1.5}$ & ${\scriptstyle-}.213\,{\scriptstyle\pm.019}$ & $9.9\,{\scriptstyle\pm1.2}$ \\
SGLD & $91.7\,{\scriptstyle\pm0.0}$ & $91.6\,{\scriptstyle\pm0.1}$ & ${\scriptstyle-}.152\,{\scriptstyle\pm.003}$ & $7.2\,{\scriptstyle\pm0.1}$ \\
DE & $91.7\,{\scriptstyle\pm0.2}$ & $91.6\,{\scriptstyle\pm0.4}$ & ${\scriptstyle-}.142\,{\scriptstyle\pm.006}$ & $7.0\,{\scriptstyle\pm0.3}$ \\
MultiSWA & $91.9\,{\scriptstyle\pm0.0}$ & $91.9\,{\scriptstyle\pm0.1}$ & ${\scriptstyle-}.133\,{\scriptstyle\pm.002}$ & $6.7\,{\scriptstyle\pm0.1}$ \\
MultiSWAG & $\textbf{\underline{91.9}}\,{\scriptstyle\pm0.0}$ & $\textbf{\underline{92.0}}\,{\scriptstyle\pm0.1}$ & ${\scriptstyle-}\textbf{\underline{.133}}\,{\scriptstyle\pm.002}$ & $\textbf{\underline{6.6}}\,{\scriptstyle\pm0.1}$ \\
\addlinespace[2pt]
\multicolumn{5}{@{}l}{\itshape CNN-GRU} \\
MAP & $90.5\,{\scriptstyle\pm0.2}$ & $90.1\,{\scriptstyle\pm0.4}$ & ${\scriptstyle-}.161\,{\scriptstyle\pm.006}$ & $8.1\,{\scriptstyle\pm0.3}$ \\
Temp. & $90.5\,{\scriptstyle\pm0.2}$ & $90.1\,{\scriptstyle\pm0.4}$ & ${\scriptstyle-}.157\,{\scriptstyle\pm.003}$ & $8.1\,{\scriptstyle\pm0.2}$ \\
SWA & $91.7\,{\scriptstyle\pm0.2}$ & $91.6\,{\scriptstyle\pm0.3}$ & ${\scriptstyle-}.135\,{\scriptstyle\pm.007}$ & $6.9\,{\scriptstyle\pm0.3}$ \\
SWAG & $91.8\,{\scriptstyle\pm0.2}$ & $91.7\,{\scriptstyle\pm0.3}$ & ${\scriptstyle-}.137\,{\scriptstyle\pm.008}$ & $6.9\,{\scriptstyle\pm0.4}$ \\
Laplace & $90.5\,{\scriptstyle\pm0.2}$ & $90.1\,{\scriptstyle\pm0.4}$ & ${\scriptstyle-}.161\,{\scriptstyle\pm.006}$ & $8.1\,{\scriptstyle\pm0.3}$ \\
VGD & $91.5\,{\scriptstyle\pm0.3}$ & $91.3\,{\scriptstyle\pm0.5}$ & ${\scriptstyle-}.148\,{\scriptstyle\pm.007}$ & $7.4\,{\scriptstyle\pm0.3}$ \\
MFVI & $90.8\,{\scriptstyle\pm0.2}$ & $90.3\,{\scriptstyle\pm0.5}$ & ${\scriptstyle-}.172\,{\scriptstyle\pm.009}$ & $8.5\,{\scriptstyle\pm0.5}$ \\
SGLD & $91.2\,{\scriptstyle\pm0.1}$ & $91.0\,{\scriptstyle\pm0.3}$ & ${\scriptstyle-}.143\,{\scriptstyle\pm.005}$ & $7.4\,{\scriptstyle\pm0.3}$ \\
DE & $91.3\,{\scriptstyle\pm0.1}$ & $91.1\,{\scriptstyle\pm0.3}$ & ${\scriptstyle-}.141\,{\scriptstyle\pm.004}$ & $7.3\,{\scriptstyle\pm0.2}$ \\
MultiSWA & $91.9\,{\scriptstyle\pm0.0}$ & $91.8\,{\scriptstyle\pm0.1}$ & ${\scriptstyle-}\underline{.133}\,{\scriptstyle\pm.001}$ & $\underline{6.8}\,{\scriptstyle\pm0.1}$ \\
MultiSWAG & $\underline{91.9}\,{\scriptstyle\pm0.0}$ & $\underline{91.8}\,{\scriptstyle\pm0.1}$ & ${\scriptstyle-}.136\,{\scriptstyle\pm.001}$ & $6.8\,{\scriptstyle\pm0.1}$ \\
\addlinespace[2pt]
\multicolumn{5}{@{}l}{\itshape Other baselines} \\
Somnotate & $\underline{90.4}$ & $\underline{91.1}$ & ${\scriptstyle-}\underline{.248}$ & $\underline{8.4}$ \\
Spindle & $89.2\,{\scriptstyle\pm0.4}$ & $89.5\,{\scriptstyle\pm0.2}$ & ${\scriptstyle-}.301\,{\scriptstyle\pm.005}$ & $10.1\,{\scriptstyle\pm0.2}$ \\
\bottomrule
\end{tabular}
\end{table}    % \label{tab:id-results}
\begin{table}[t]
\centering
\caption{\textbf{Out-of-distribution performance}, each metric averaged over the five OOD cohorts (NT1 + SleepyRat A--D), computed per training seed; mean $\pm$ 95\% CI over training seeds ($t$-based); for DE, MultiSWA and MultiSWAG, mean $\pm$ half the 2.5--97.5\% interval over 100 bootstrap draws of ten members of that average. \textbf{bold} = best in the column over all models, \underline{underline} = best within its section.}
\label{tab:ood-results}
\scriptsize
\setlength{\tabcolsep}{4pt}
\begin{tabular}{lcccc}
\toprule
Model & MF1\,$\uparrow$ & $\kappa$\,$\uparrow$ & LPD\,$\uparrow$ & Brier\,$\downarrow$ \\
\midrule
\multicolumn{5}{@{}l}{\itshape U-Sleep} \\
MAP & $76.0\,{\scriptstyle\pm4.4}$ & $72.1\,{\scriptstyle\pm6.3}$ & ${\scriptstyle-}.481\,{\scriptstyle\pm.149}$ & $24.2\,{\scriptstyle\pm6.7}$ \\
Temp. & $76.0\,{\scriptstyle\pm4.4}$ & $72.1\,{\scriptstyle\pm6.3}$ & ${\scriptstyle-}.460\,{\scriptstyle\pm.122}$ & $23.9\,{\scriptstyle\pm6.3}$ \\
SWA & $78.0\,{\scriptstyle\pm3.2}$ & $74.3\,{\scriptstyle\pm5.9}$ & ${\scriptstyle-}.435\,{\scriptstyle\pm.105}$ & $21.7\,{\scriptstyle\pm4.8}$ \\
SWAG & $79.2\,{\scriptstyle\pm2.8}$ & $75.7\,{\scriptstyle\pm5.3}$ & ${\scriptstyle-}.376\,{\scriptstyle\pm.060}$ & $19.8\,{\scriptstyle\pm3.6}$ \\
Laplace & $76.0\,{\scriptstyle\pm4.4}$ & $72.1\,{\scriptstyle\pm6.3}$ & ${\scriptstyle-}.481\,{\scriptstyle\pm.149}$ & $24.2\,{\scriptstyle\pm6.7}$ \\
VGD & $75.2\,{\scriptstyle\pm6.2}$ & $68.2\,{\scriptstyle\pm12.0}$ & ${\scriptstyle-}.473\,{\scriptstyle\pm.202}$ & $25.7\,{\scriptstyle\pm10.6}$ \\
MFVI & $79.6\,{\scriptstyle\pm2.0}$ & $73.2\,{\scriptstyle\pm3.5}$ & ${\scriptstyle-}.415\,{\scriptstyle\pm.038}$ & $22.6\,{\scriptstyle\pm2.5}$ \\
SGLD & $80.6\,{\scriptstyle\pm1.3}$ & $76.1\,{\scriptstyle\pm2.0}$ & ${\scriptstyle-}.362\,{\scriptstyle\pm.028}$ & $19.4\,{\scriptstyle\pm1.5}$ \\
DE & $82.1\,{\scriptstyle\pm1.5}$ & $\underline{79.5}\,{\scriptstyle\pm2.4}$ & ${\scriptstyle-}\textbf{\underline{.299}}\,{\scriptstyle\pm.019}$ & $\textbf{\underline{16.2}}\,{\scriptstyle\pm1.3}$ \\
MultiSWA & $82.2\,{\scriptstyle\pm0.9}$ & $79.3\,{\scriptstyle\pm2.0}$ & ${\scriptstyle-}.315\,{\scriptstyle\pm.018}$ & $16.6\,{\scriptstyle\pm1.2}$ \\
MultiSWAG & $\textbf{\underline{82.5}}\,{\scriptstyle\pm0.8}$ & $79.4\,{\scriptstyle\pm1.8}$ & ${\scriptstyle-}.311\,{\scriptstyle\pm.015}$ & $16.5\,{\scriptstyle\pm0.9}$ \\
\addlinespace[2pt]
\multicolumn{5}{@{}l}{\itshape CNN-GRU} \\
MAP & $77.1\,{\scriptstyle\pm3.2}$ & $73.8\,{\scriptstyle\pm2.1}$ & ${\scriptstyle-}.423\,{\scriptstyle\pm.064}$ & $21.5\,{\scriptstyle\pm2.2}$ \\
Temp. & $77.1\,{\scriptstyle\pm3.2}$ & $73.8\,{\scriptstyle\pm2.1}$ & ${\scriptstyle-}.396\,{\scriptstyle\pm.053}$ & $21.0\,{\scriptstyle\pm2.2}$ \\
SWA & $77.2\,{\scriptstyle\pm2.7}$ & $72.8\,{\scriptstyle\pm2.9}$ & ${\scriptstyle-}.396\,{\scriptstyle\pm.048}$ & $21.2\,{\scriptstyle\pm2.4}$ \\
SWAG & $78.8\,{\scriptstyle\pm2.2}$ & $74.6\,{\scriptstyle\pm2.6}$ & ${\scriptstyle-}.361\,{\scriptstyle\pm.037}$ & $19.5\,{\scriptstyle\pm2.1}$ \\
Laplace & $77.1\,{\scriptstyle\pm3.2}$ & $73.8\,{\scriptstyle\pm2.1}$ & ${\scriptstyle-}.423\,{\scriptstyle\pm.064}$ & $21.5\,{\scriptstyle\pm2.2}$ \\
VGD & $79.2\,{\scriptstyle\pm1.6}$ & $76.3\,{\scriptstyle\pm0.4}$ & ${\scriptstyle-}.356\,{\scriptstyle\pm.055}$ & $18.8\,{\scriptstyle\pm1.8}$ \\
MFVI & $77.4\,{\scriptstyle\pm3.4}$ & $75.1\,{\scriptstyle\pm0.8}$ & ${\scriptstyle-}.390\,{\scriptstyle\pm.016}$ & $20.2\,{\scriptstyle\pm0.4}$ \\
SGLD & $78.7\,{\scriptstyle\pm1.2}$ & $74.9\,{\scriptstyle\pm0.7}$ & ${\scriptstyle-}.375\,{\scriptstyle\pm.036}$ & $19.7\,{\scriptstyle\pm0.9}$ \\
DE & $\underline{79.7}\,{\scriptstyle\pm0.7}$ & $\underline{76.3}\,{\scriptstyle\pm0.3}$ & ${\scriptstyle-}\underline{.347}\,{\scriptstyle\pm.018}$ & $\underline{18.7}\,{\scriptstyle\pm0.5}$ \\
MultiSWA & $77.7\,{\scriptstyle\pm0.5}$ & $72.7\,{\scriptstyle\pm0.5}$ & ${\scriptstyle-}.382\,{\scriptstyle\pm.009}$ & $20.9\,{\scriptstyle\pm0.5}$ \\
MultiSWAG & $78.9\,{\scriptstyle\pm0.5}$ & $74.4\,{\scriptstyle\pm0.4}$ & ${\scriptstyle-}.352\,{\scriptstyle\pm.007}$ & $19.3\,{\scriptstyle\pm0.4}$ \\
\addlinespace[2pt]
\multicolumn{5}{@{}l}{\itshape Other baselines} \\
Somnotate & $72.1$ & $68.7$ & ${\scriptstyle-}.863$ & $32.0$ \\
Spindle & $\underline{79.6}\,{\scriptstyle\pm1.2}$ & $\textbf{\underline{79.7}}\,{\scriptstyle\pm1.3}$ & ${\scriptstyle-}\underline{.676}\,{\scriptstyle\pm.034}$ & $\underline{19.9}\,{\scriptstyle\pm1.4}$ \\
\bottomrule
\end{tabular}
\end{table}       % \label{tab:ood-results}

\paragraph{Training, model selection and evaluation}
Using i.i.d. Gaussian priors, the MAP models were trained using Adam ($\beta_1=0.9$, $\beta_2=0.999$), a learning rate of $10^{-4}$, and a batch size of 64 input sequences. Both models were trained for a fixed budget of approximately 57 full training epochs.

For each backbone, we selected the prior variance that achieved the highest validation macro F1 (MF1) (see Appendix~\ref{app:map-prior-selection} for further details). The selected prior was then held fixed across all posterior approximation methods in Table~\ref{tab:posteriors}. 

MF1 was the primary classification metric, supplemented by Cohen's $\kappa$. Probabilistic performance was evaluated using log predictive density (LPD) and Brier score. All metrics were computed using the predictive distribution in Eq.~\eqref{eq:predictive}. Single-model methods were evaluated across five random seeds, and we report the mean and $95\%$ confidence interval across runs. DE, MultiSWA, and MultiSWAG used $M=10$ independently initialized models, with $95\%$ empirical intervals obtained from 100 random ensembles sampled with replacement from a pool of 20 models. These intervals likely underestimate variability relative to independent ensembles. Further details on tuning and evaluation are provided in Appendix~\ref{app:bayes}.

Lastly, we evaluated predictive entropy for detecting OOD recordings, inter-scorer disagreement, and artifacts, using the area under the receiver operating characteristic curve (AUROC). OOD detection used per-recording mean entropy to distinguish WT test recordings from OOD recordings, whereas disagreement and artifact detection used window-level entropy and the original \textsc{spindle} annotations. We evaluated selective prediction using the area under the risk-coverage curve (AURC) \citep{Geifman.etal-2019}.

\subsection{MAP models are strong baselines}
\label{sec:exp-baselines}

We first compared the U-Sleep and CNN-GRU MAP models with \textit{Somnotate} and \textit{Spindle} on the ID WT test set. Table~\ref{tab:id-results} shows strong and comparable classification performance across all four models: MF1 ranges from $89.2$ to $90.5$ and
Cohen's $\kappa$ from $88.6$ to $91.1$, with CNN-GRU MAP achieving the highest mean MF1 and Somnotate the highest $\kappa$. The MAP models achieve substantially higher LPD than both \textit{Somnotate} and \textit{Spindle}; CNN-GRU MAP also obtains the lowest Brier score, while U-Sleep MAP lies between Somnotate and Spindle. Thus, the MAP models are competitive with established sleep-scoring models and provide strong deterministic baselines for the following comparison of posterior approximation methods.

\begin{figure}[t]
    \centering
    \includegraphics[width=\columnwidth]{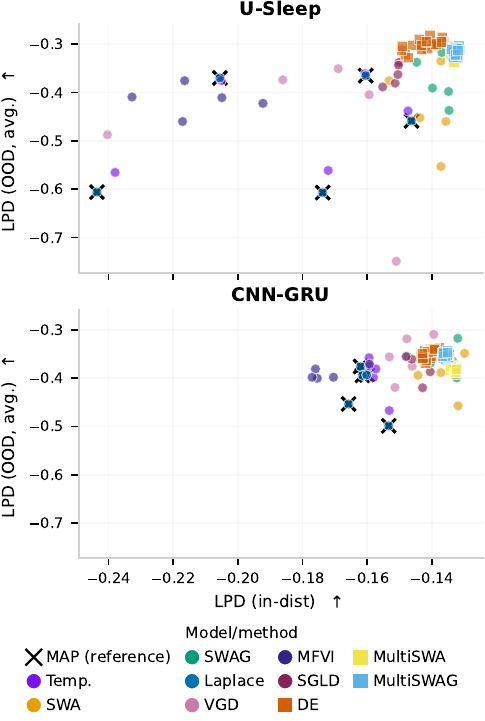}
    \caption{
    ID versus mean OOD LPD for U-Sleep and CNN-GRU
    (higher is better) across seeds. Black crosses denote the corresponding
    MAP models, and each dot represents a seed.
    }
    \label{fig:id-ood-lpd}
\end{figure}

\begin{table*}[ht]
\centering
\caption{\textbf{Average LPD rank.} Models were ranked jointly within each cohort, and ranks were averaged over cohorts and backbones ($1=$ best; lower is better). Other baselines are ranked individually. \textbf{Bold} and \underline{underlining} denote the best overall and within each model group, respectively.}
\label{tab:rank-averaged-transposed-lpd}
\scriptsize
\setlength{\tabcolsep}{3pt}
\begin{tabular*}{\textwidth}{@{\extracolsep{\fill}}lccccccccccccc@{}}
\toprule
 & \multicolumn{8}{c}{Single solution/trajectory} & \multicolumn{3}{c}{Multi-basin} & \multicolumn{2}{c}{Other baselines} \\
\cmidrule(lr){2-9} \cmidrule(lr){10-12} \cmidrule(lr){13-14}
Scope & MAP & Temp. & SWA & SWAG & Laplace & VGD & MFVI & SGLD & DE & MultiSWA & MultiSWAG & Somnotate & Spindle \\
\midrule
ID & 17.5 & 16.5 & 6.5 & 6.5 & 18.5 & 15.0 & 19.5 & 12.0 & 9.0 & \textbf{2.5} & \underline{3.0} & 23.0 & 24.0 \\
OOD & 18.1 & 16.0 & 13.0 & 9.1 & 18.5 & 12.6 & 13.5 & 9.9 & \underline{6.2} & 6.8 & \textbf{4.7} & 23.6 & 19.6 \\
ID + OOD & 18.0 & 16.1 & 11.9 & 8.7 & 18.5 & 13.0 & 14.5 & 10.2 & 6.7 & \underline{6.1} & \textbf{4.4} & 23.5 & 20.3 \\
\bottomrule
\end{tabular*}
\end{table*}

\subsection{BMA yields modest in-distribution gains}
\label{sec:exp-id-inference}

Next, we compared the approximate BMA methods on the ID WT test set. Table~\ref{tab:id-results} shows that most methods improve classification and probabilistic performance on the ID WT test set, but the gains are modest. The best results increase MF1 by $2.2$ percentage points for U-Sleep and $1.4$ points for CNN-GRU, with corresponding improvements in LPD and Brier score. SWA, SWAG, and the multi-basin methods perform strongest, although the differences among them are small. LLLA matches MAP to the reported precision, MFVI slightly improves classification but worsens probabilistic performance, and temperature scaling provides only marginal probabilistic gains. 

\subsection{BMA generally improves robustness under natural distribution shift}
\label{sec:exp-ood}

Next, we evaluated all models on the five OOD cohorts. Table~\ref{tab:ood-results} shows that distribution shift on average degrades both classification accuracy and calibration, while increasing the differences between inference methods. Most BMA methods improve on their corresponding MAP model for OOD data, with larger gains than ID. For U-Sleep, MultiSWAG increases MF1 from $76.0$ to $82.5$, while the DE improves LPD from $-0.481$ to $-0.299$ and reduces the Brier score from $24.2$ to $16.2$. For CNN-GRU, the DE increases MF1 from $77.1$ to $79.7$, improves LPD from $-0.423$ to $-0.347$, and reduces the Brier score from $21.5$ to $18.7$. Temperature scaling yields only marginally better OOD calibration than MAP, indicating that the BMA gains cannot be attributed to a global rescaling of the MAP logits, mirroring the findings of \citet{Ovadia.etal-2019}.

Across both backbones, multi-basin methods are the strongest overall, which is consistent with the literature \citep{Wilson.Izmailov-2020, Jordahn.etal-2025}. The DE achieves the best OOD LPD and Brier score for both backbones, while MultiSWAG achieves the highest overall MF1. Among the single-basin methods, SWAG and SGLD perform well across architectures, and SWAG outperforms SWA on every mean OOD metric, as does MultiSWAG relative to MultiSWA. The effects nevertheless remain architecture dependent: VGD is competitive for CNN-GRU but not U-Sleep, while LLLA reproduces the corresponding MAP results. Among the literature baselines, Spindle retains strong classification performance but substantially worse LPD, whereas Somnotate degrades the most. 
The average ranks in Table~\ref{tab:rank-averaged-transposed-lpd} confirm MultiSWAG as the strongest method across backbones in terms of LPD by a clear margin, with an average rank of 4.4 across ID and OOD datasets, compared with 6.1 for MultiSWA and 6.7 for the DE. MultiSWAG also ranks first on MF1 alone, although by a smaller margin (Appendix Table~\ref{tab:rank-averaged-transposed-mf1}).

\begin{table}[t]
\centering
\caption{\textbf{Predictive entropy for detection and selective prediction.} AUROC (higher is better) for detecting OOD recordings (per-recording mean entropy, WT vs.\ the five held-out OOD cohorts), artifact epochs, and inter-rater disagreement (epoch-level entropy, \textsc{spindle} cohorts); AURC (lower is better) for rejecting the most uncertain epochs on the OOD cohorts. Mean $\pm$ 95\% interval over training seeds; for DE, MultiSWA, and MultiSWAG, the 2.5--97.5\% interval over 100 resampled ensembles. \textbf{Bold}: best overall; \underline{underline}: best within the model group.}
\label{tab:uncertainty-utility}
\scriptsize
\setlength{\tabcolsep}{4pt}
\begin{tabular}{lcccc}
\toprule
 & \multicolumn{3}{c}{AUROC $\uparrow$} & \multicolumn{1}{c}{AURC $\downarrow$} \\
\cmidrule(lr){2-4} \cmidrule(lr){5-5}
Model & OOD & Artifact & Disagree & Reject \\
\midrule
\multicolumn{5}{@{}l}{\itshape U-Sleep} \\
MAP & $.72\,{\scriptstyle\pm.06}$ & $.44\,{\scriptstyle\pm.03}$ & $.76\,{\scriptstyle\pm.02}$ & $.054\,{\scriptstyle\pm.015}$ \\
Temp. & $.71\,{\scriptstyle\pm.06}$ & $.44\,{\scriptstyle\pm.03}$ & $.76\,{\scriptstyle\pm.02}$ & $.054\,{\scriptstyle\pm.015}$ \\
SWA & $.80\,{\scriptstyle\pm.03}$ & $.49\,{\scriptstyle\pm.03}$ & $.76\,{\scriptstyle\pm.01}$ & $.049\,{\scriptstyle\pm.019}$ \\
SWAG & $.82\,{\scriptstyle\pm.02}$ & $.51\,{\scriptstyle\pm.03}$ & $.75\,{\scriptstyle\pm.01}$ & $.041\,{\scriptstyle\pm.014}$ \\
Laplace & $.72\,{\scriptstyle\pm.06}$ & $.44\,{\scriptstyle\pm.03}$ & $.76\,{\scriptstyle\pm.02}$ & $.054\,{\scriptstyle\pm.015}$ \\
VGD & $.72\,{\scriptstyle\pm.07}$ & $.59\,{\scriptstyle\pm.08}$ & $.72\,{\scriptstyle\pm.05}$ & $.076\,{\scriptstyle\pm.069}$ \\
MFVI & $.71\,{\scriptstyle\pm.07}$ & $.65\,{\scriptstyle\pm.03}$ & $.71\,{\scriptstyle\pm.01}$ & $.052\,{\scriptstyle\pm.013}$ \\
SGLD & $.78\,{\scriptstyle\pm.02}$ & $\underline{.66}\,{\scriptstyle\pm.02}$ & $.73\,{\scriptstyle\pm.01}$ & $.041\,{\scriptstyle\pm.005}$ \\
DE & $.81\,{\scriptstyle\pm.03}$ & $.51\,{\scriptstyle\pm.05}$ & $\underline{.77}\,{\scriptstyle\pm.02}$ & $.027\,{\scriptstyle\pm.005}$ \\
MultiSWA & $.83\,{\scriptstyle\pm.01}$ & $.53\,{\scriptstyle\pm.03}$ & $.76\,{\scriptstyle\pm.01}$ & $.028\,{\scriptstyle\pm.004}$ \\
MultiSWAG & $\underline{.83}\,{\scriptstyle\pm.01}$ & $.56\,{\scriptstyle\pm.03}$ & $.76\,{\scriptstyle\pm.01}$ & $\mathbf{\underline{.026}}\,{\scriptstyle\pm.004}$ \\
\addlinespace[2pt]
\multicolumn{5}{@{}l}{\itshape CNN-GRU} \\
MAP & $.75\,{\scriptstyle\pm.04}$ & $.54\,{\scriptstyle\pm.04}$ & $.75\,{\scriptstyle\pm.02}$ & $.055\,{\scriptstyle\pm.017}$ \\
Temp. & $.77\,{\scriptstyle\pm.04}$ & $.54\,{\scriptstyle\pm.04}$ & $.75\,{\scriptstyle\pm.02}$ & $.055\,{\scriptstyle\pm.017}$ \\
SWA & $.84\,{\scriptstyle\pm.03}$ & $.70\,{\scriptstyle\pm.03}$ & $.72\,{\scriptstyle\pm.02}$ & $.064\,{\scriptstyle\pm.025}$ \\
SWAG & $.83\,{\scriptstyle\pm.02}$ & $.69\,{\scriptstyle\pm.03}$ & $.74\,{\scriptstyle\pm.02}$ & $.052\,{\scriptstyle\pm.015}$ \\
Laplace & $.75\,{\scriptstyle\pm.04}$ & $.54\,{\scriptstyle\pm.04}$ & $.75\,{\scriptstyle\pm.02}$ & $.055\,{\scriptstyle\pm.017}$ \\
VGD & $.81\,{\scriptstyle\pm.02}$ & $.59\,{\scriptstyle\pm.06}$ & $.77\,{\scriptstyle\pm.01}$ & $.049\,{\scriptstyle\pm.026}$ \\
MFVI & $.73\,{\scriptstyle\pm.05}$ & $.50\,{\scriptstyle\pm.05}$ & $\mathbf{\underline{.77}}\,{\scriptstyle\pm.01}$ & $.050\,{\scriptstyle\pm.009}$ \\
SGLD & $.76\,{\scriptstyle\pm.05}$ & $.54\,{\scriptstyle\pm.04}$ & $.76\,{\scriptstyle\pm.01}$ & $.042\,{\scriptstyle\pm.013}$ \\
DE & $.81\,{\scriptstyle\pm.02}$ & $.57\,{\scriptstyle\pm.03}$ & $.76\,{\scriptstyle\pm.01}$ & $\underline{.036}\,{\scriptstyle\pm.006}$ \\
MultiSWA & $\underline{.85}\,{\scriptstyle\pm.01}$ & $\mathbf{\underline{.71}}\,{\scriptstyle\pm.01}$ & $.73\,{\scriptstyle\pm.01}$ & $.055\,{\scriptstyle\pm.006}$ \\
MultiSWAG & $.84\,{\scriptstyle\pm.01}$ & $.69\,{\scriptstyle\pm.01}$ & $.74\,{\scriptstyle\pm.01}$ & $.048\,{\scriptstyle\pm.004}$ \\
\addlinespace[2pt]
\multicolumn{5}{@{}l}{\itshape Other baselines} \\
Somnotate & $.86$ & $\underline{.44}$ & $.71$ & $.072$ \\
Spindle & $\mathbf{\underline{.88}}\,{\scriptstyle\pm.01}$ & $.40\,{\scriptstyle\pm.04}$ & $\underline{.76}\,{\scriptstyle\pm.01}$ & $\underline{.043}\,{\scriptstyle\pm.003}$ \\
\bottomrule
\end{tabular}
\end{table}

Figure~\ref{fig:id-ood-lpd} illustrates the same pattern across seeds. Most methods move above and to the right of their MAP reference, indicating improved LPD both ID and OOD, with larger OOD gains for U-Sleep. Multi-basin methods occupy the upper-right region, but their small spread should be interpreted cautiously because the sampled ensembles overlap in their members, as described in Section~\ref{sec:exp-setup}. Cohort-specific reliability diagrams in Appendix Figure~\ref{fig:reliability-ood-cohorts} show substantial variation across OOD cohorts. BMA often reduces pronounced MAP overconfidence, but the effect is not uniform and some methods become underconfident. 

Overall, these results support previous findings that multi-basin methods are robust under distribution shift \citep{Ovadia.etal-2019,Wilson.Izmailov-2020,Seligmann.etal-2023}. MultiSWAG does not consistently outperform the DE, suggesting that much of the benefit comes from between-basin averaging \citep{Jordahn.etal-2025}. However, SWAG outperforms SWA and MultiSWAG outperforms MultiSWA on OOD data, while MultiSWAG ranks best overall on both MF1 and LPD, suggesting an additional benefit from local posterior structure \citep{Maddox.etal-2019,Wilson.Izmailov-2020}.

\subsection{Uncertainty has task-dependent utility}
\label{sec:exp-utility}

Finally, we evaluated predictive entropy for detecting OOD recordings, artifacts, and inter-rater disagreement, and for selective prediction. Table~\ref{tab:uncertainty-utility} reports the detection AUROCs and rejection AURC; mean entropy values are provided in Appendix Table~\ref{tab:uncertainty-utility-full}. Most BMA methods improve OOD detection relative to MAP, consistent with previous work \citep{Jordahn.etal-2025,Wilson.Izmailov-2020}. For U-Sleep, the best AUROC increases from $.72$ for MAP to $.83$ for MultiSWA and MultiSWAG, while for CNN-GRU it increases from $.75$ to $.85$ for MultiSWA. Spindle and Somnotate also perform well, with AUROCs of $.88$ and $.86$, respectively. For artifact detection, the best AUROC increases from $.44$ to $.66$ for U-Sleep for SGLD, and from $.54$ to $.71$ for CNN-GRU, achieved by MultiSWA. In contrast, disagreement detection shows no systematic improvement: AUROCs remain between $.71$ and $.77$, and several BMA methods match or underperform MAP. Finally, selective prediction improves for some, but not all, methods. MultiSWAG reduces OOD AURC from $.054$ to $.026$ for U-Sleep, while the CNN-GRU DE reduces it from $.055$ to $.036$.

Overall, predictive entropy has task-dependent utility. BMA generally improves OOD detection, often improves artifact detection and selective prediction, but does not systematically improve disagreement detection. One possible explanation is that predictive entropy combines different sources of uncertainty, whereas scorer disagreement may more directly reflect ambiguity in the target or annotation process \citep{vanGorp.etal-2022,Bakker.etal-2023}. We therefore additionally decomposed predictive entropy into aleatoric and epistemic components, but found no consistent improvement in disagreement detection: aleatoric uncertainty did not consistently outperform total predictive entropy, while epistemic uncertainty was generally less informative (Appendix Table~\ref{tab:uncertainty-decomposition}). 

\section{DISCUSSION \& SUMMARY}

Across two neural network architectures and five shifted datasets, approximate BMA provides modest ID gains but substantially larger improvements under distribution shift. Multi-basin methods are consistently more robust under shift across both classification and probabilistic metrics, extending previous findings to biomedical time-series classification \citep{Ovadia.etal-2019,Wilson.Izmailov-2020,Seligmann.etal-2023}. However, the utility of predictive uncertainty is similarly task-dependent, and no single method dominates across all metrics and architectures. BMA generally improves OOD detection and often artifact detection and selective prediction, but does not systematically improve inter-rater disagreement detection. Thus, improved predictive probabilities do not necessarily translate into uncertainty estimates that are equally informative for every downstream task. From a practical perspective, the strongest multi-basin methods require several independently trained models, whereas SWAG provides a single-trajectory alternative with consistent OOD gains across both architectures \citep{Maddox.etal-2019}. Finally, while multi-modal BMA are found to generally improve both accuracy and calibration, one has to balance the increase in predictive performance against the significant increase in computational cost.

\paragraph{Limitations and future work}
Our conclusions are limited to rodent sleep scoring and five relatively small OOD cohorts. In addition, because several sources of distribution shift co-occur within cohorts, we cannot reliably isolate the effects of individual shift factors, as can be done in controlled simulations. Future work should investigate whether our findings generalize to additional biomedical time-series tasks and distribution shifts.

Overall, approximate BMA, particularly multi-basin methods, improves the robustness of biomedical time-series classifiers under distribution shift, but reliable deployment also requires uncertainty estimates to be validated for the specific decisions they are intended to support.

\subsection*{AI use statement}
In this work, we used generative AI tools for code generation, visualization, and to refine writing. We have not used generative AI tools for generating research questions and experimental design. We have reviewed all AI-assisted work. For example, LLM-generated code was verified and tested using both manual and automated unit tests. We take responsibility for the final content of this work, including text, claims, or artifacts produced with the aid of generative AI.

\bibliography{references}

\onecolumn
\appendix

\begin{center}
 {\LARGE\bfseries Supplementary Materials}
\end{center}
\vspace{1em}

\section{Experimental setup}
\label{app:exp-setup}
\subsection{Datasets}
\label{app:data}

Table~\ref{tab:datasets-full} summarizes the main characteristics of all datasets used in this paper, including the source of the dataset, the types of electrophysiological signals, and the shift. The WT and NT1 dataset has been collected at H. Lundbeck A/S across multiple research studies. 

\paragraph{Signals and preprocessing}
For the WT and NT1 cohorts, we used one electrocorticographic (ECoG) channel, one depth local-field-potential (LFP) channel, and one electromyographic (EMG) channel (Table~\ref{tab:datasets-full}). For the \textsc{spindle} cohorts, we used two ECoG channels and one EMG channel. Consequently, the second neural input channel represents LFP in the WT and NT1 cohorts but ECoG in the \textsc{spindle} cohorts, introducing an additional shift in signal type.

For both neural network backbones, all channels were resampled to $128$~Hz, centered by their median, scaled by their interquartile range, and clipped to $[-20,20]$. The literature baselines retained their method-specific preprocessing procedures, described in Appendices~\ref{app:spindle} and~\ref{app:somnotate}.

\paragraph{Annotations}
The WT and NT1 datasets were annotated by a single scorer using $\mathcal{Y}=\{\text{wake},\text{NREM},\text{REM}\}$. In contrast, the \textsc{spindle} cohorts were annotated independently by two scorers using the expanded label set
$$
\mathcal{Y}_{\textsc{spindle}}
=
\mathcal{Y}
\cup
\{
A_{\text{wake}},
A_{\text{NREM}},
A_{\text{REM}}
\},
$$
where $A_{\text{wake}}$, $A_{\text{NREM}}$, and $A_{\text{REM}}$ denote artifact-containing windows assigned to the corresponding sleep stage. 

\subsection{Model training, selection, and evaluation}

\paragraph{Model training}
Training schedules were implemented in blocks of $10^5$ processed labels, with approximately $3.5$ million labels in the full training set. The MAP models were trained for 2000 blocks, corresponding to approximately 57 passes through the training set. VGD, MFVI, and SGLD used 3000 blocks, corresponding to approximately 85 passes, while SWA and SWAG used five transition blocks followed by 200 collection blocks after initialization from the MAP solution.

\begin{figure*}[t]
    \centering
    \includegraphics[width=\textwidth]
    {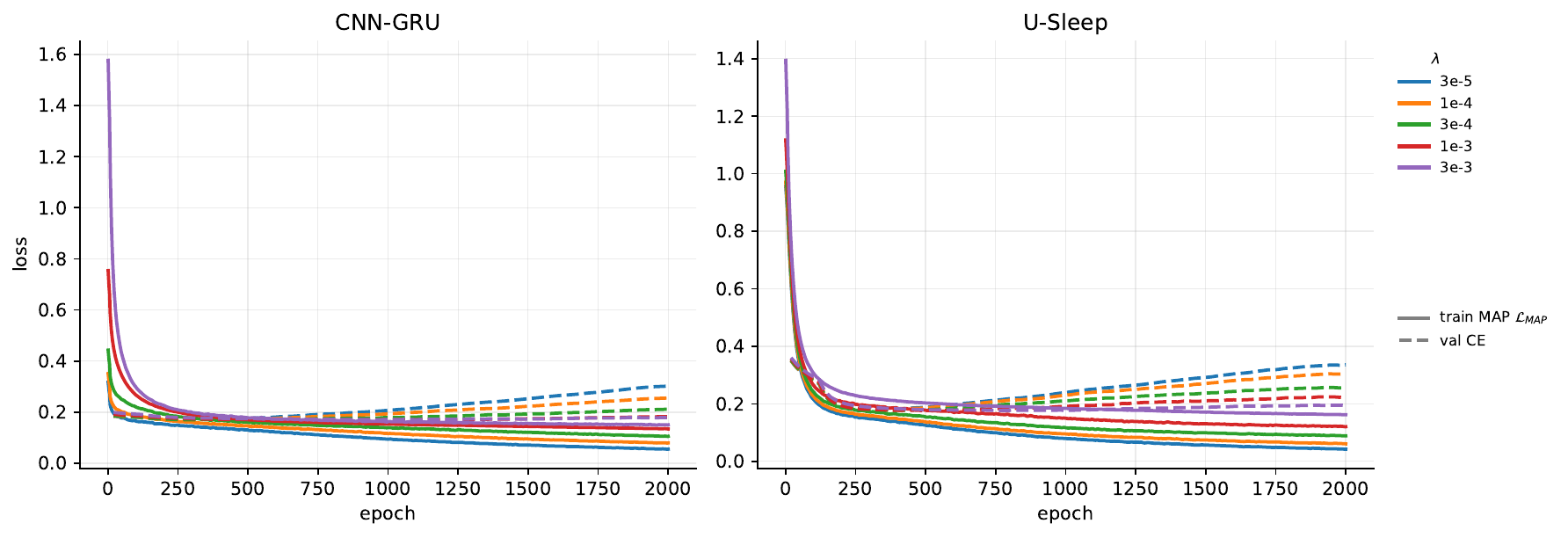}
    \caption{
    MAP optimization trajectories across prior strengths for CNN-GRU and
    U-Sleep. Solid lines show the training MAP objective and dashed lines show
    validation cross-entropy, averaged across three random seeds. All
    configurations were optimized for 2000 blocks (approximately 57 passes through the training set). The trajectories were used to verify convergence. The backbone-specific prior strengths were selected using mean validation MF1.
    }
    \label{fig:map-optimization}
\end{figure*}

We use this block notation below whenever the exact optimization or collection schedule is relevant. The same unit was used for the loss curves in Figure~\ref{fig:map-optimization} and to express the early-stopping patience for the \textsc{spindle} model (Appendix Section~\ref{app:spindle}) and MAP model-complexity tuning (Appendix Section~\ref{app:determin-backbones}).

\paragraph{Model selection}
Hyperparameter tuning was based on mean validation MF1. This is visualized in Figure~\ref{fig:baseline-sizes} for the MAP model-complexity tuning and in Figure~\ref{fig:bayes-sensitivity} for the MAP prior variance and posterior approximation hyperparameters. For each method, we tuned one main method-specific hyperparameter while holding all other settings fixed.

\paragraph{Posterior inference}
For sampling-based posterior approximations, we used $S=30$ Monte Carlo samples, as summarized in Table~\ref{tab:bayes-hyperparameters}. DE and MultiSWA used $M=10$ independently trained components, while MultiSWAG used $M=10$ components with $S=30$ samples per component.

\paragraph{Evaluation}
In addition to the metrics reported in the main tables, we report accuracy, expected calibration error (ECE), and predictive entropy ($H$) in Tables~\ref{tab:id-results-wide} and~\ref{tab:ood-results-wide}. For OOD, artifact, and inter-rater disagreement detection, as well as the reject option in Section~\ref{sec:exp-utility}, we used predictive entropy as the uncertainty score, with the main results reported in Table~\ref{tab:uncertainty-utility}. We further decomposed predictive entropy into aleatoric uncertainty (AU) and epistemic uncertainty (EU), as reported in Appendix Table~\ref{tab:uncertainty-decomposition}.

For output window $w$ of a new sequence $\mathbf{X}^*$, total uncertainty (TU) is the predictive entropy
$
\mathrm{TU}
=
H\!\left[
p(y_w^*\mid\mathbf{X}^*,\mathcal{D})
\right].
$ The aleatoric component is the expected entropy of the component predictions, approximated using the same $S$ predictive components as in Eq.~\eqref{eq:predictive},
$$
\mathrm{AU}
=
\mathbb{E}_{q_m(\boldsymbol{\theta})}
\left[
H\!\left[
p(y_w^*\mid\mathbf{X}^*,\boldsymbol{\theta})
\right]
\right]
\approx
\frac{1}{S}
\sum_{s=1}^{S}
H\!\left[
p(y_w^*\mid\mathbf{X}^*,\boldsymbol{\theta}^{(s)})
\right],
$$
where $\boldsymbol{\theta}^{(s)}\sim q_m(\boldsymbol{\theta})$. The epistemic component is the remaining uncertainty,
$$
\mathrm{EU}
=
\mathrm{TU}-\mathrm{AU}
=
I(y_w^*,\boldsymbol{\theta}\mid\mathbf{X}^*,\mathcal{D}).
$$
For single-basin methods, the components were the Monte Carlo predictive samples used to form the averaged prediction. For DE and MultiSWA, the components were the individual ensemble members, while for MultiSWAG they comprised posterior samples across all ensemble members. Point-estimate methods correspond to the $S=1$ case, for which $\mathrm{AU}=\mathrm{TU}$ and $\mathrm{EU}=0$. All quantities were computed per output window using natural logarithms, with probabilities clipped at $10^{-12}$ for numerical stability.

\paragraph{Temperature scaling}
Temperature scaling \citep{Guo.etal-2017} keeps the MAP parameters fixed and fits a single scalar $\gamma>0$ that rescales the logits before the softmax, $p(y_w^*\mid\mathbf{X}^*)=\operatorname{softmax}(f_{\boldsymbol{\theta}}(\mathbf{X}^*)_w/\gamma)$, so that hard predictions are unchanged and only the confidence is adjusted. For each backbone and random seed, $\gamma$ was fitted to the final MAP checkpoint by minimizing the mean cross-entropy of the rescaled probabilities over all windows of the WT validation set, optimizing $\log\gamma$ with L-BFGS from $\gamma=1$. The fitted temperature was then held fixed for all test and OOD datasets.

\paragraph{Computing infrastructure}
All experiments were run on Amazon Web Services (AWS) EC2 instances. Depending on the setup, we utilized either \texttt{g4dn.4xlarge} instances equipped with an NVIDIA T4 GPU (16~GB VRAM) or \texttt{g5.4xlarge} instances equipped with an NVIDIA A10G GPU (24~GB VRAM). Independent of the GPU type, both configurations provided identical environments with 16 virtual CPU cores and 64~GB of RAM.

\section{Baseline models}
\label{app:baselines}

This section describes the implementation and hyperparameter tuning of the Somnotate, Spindle, U-Sleep, and CNN-GRU baselines. Table~\ref{tab:baselines} summarizes the main characteristics of the models.

\begin{table*}[t]
\centering
\caption{Datasets used for training, model selection, and evaluation, grouped by training/validation and test sets. $N_{\mathrm{rec}}$ denotes the number of 24\,h recordings. ID denotes in-distribution, whereas OOD denotes out-of-distribution.}
\label{tab:datasets-full}
\small
\setlength{\tabcolsep}{5pt}
\begin{tabular}{@{}llllrll@{}}
\toprule
Dataset & Species/genotype & Source & Channels & $N_{\mathrm{rec}}$ & Role & Shift w.r.t.\ training \\
\midrule
\multicolumn{7}{@{}l}{\textit{Training and validation}} \\
WT  & Mouse/wild-type  & Lundbeck & ECoG, LFP, EMG & 169 & Train      & --- \\
WT  & Mouse/wild-type  & Lundbeck  & ECoG, LFP, EMG & 36  & Validation & None \\
\addlinespace[2pt]
\multicolumn{7}{@{}l}{\textit{Test}} \\
WT  & Mouse/wild-type  & Lundbeck  & ECoG, LFP, EMG & 36  & Test (ID)  & None \\
NT1 & Mouse/NT1       & Lundbeck & ECoG, LFP, EMG & 28 & Test (OOD) & Genotype \\
\textsc{spindle}-A & Mouse/wild-type     & BrownLab   & $2\times$ECoG, EMG & 4 & Test (OOD) & Lab \\
\textsc{spindle}-D & Mouse/wild-type     & TidisLab   & $2\times$ECoG, EMG & 6 & Test (OOD) & Lab \\
\textsc{spindle}-B & Mouse/mutant & BrownLab   & $2\times$ECoG, EMG & 4 & Test (OOD) & Lab $+$ genotype \\
\textsc{spindle}-C & Rat/wild-type       & BaumannLab & $2\times$ECoG, EMG & 8 & Test (OOD) & Lab $+$ species \\
\bottomrule
\end{tabular}
\end{table*}

\subsection{Somnotate}
\label{app:somnotate}
We utilized Somnotate using the authors' published code \citep{Brodersen.etal-2024}, which is dual-licensed under GPLv3 for academic use. Somnotate combines LDA with a three-state Gaussian HMM applied to multitaper spectrogram features.

\paragraph{Spectrogram construction}
We evaluated the 1-s temporal resolution used by \citet{Brodersen.etal-2024} and a 4-s resolution matching the annotations in this study. For the 1-s setting, each 4-s annotation was expanded to four 1-s intervals, and the four corresponding posterior state-probability vectors were averaged for validation. The 4-s setting achieved the highest validation MF1 and was used for the final experiments, yielding one prediction and probability vector per annotated window.

Spectral features were computed using the authors' multitaper implementation, followed by the $\log(P+1)$ transformation and per-recording standardization. To obtain a common representation for recordings sampled at 128 and 256\,Hz, we retained frequencies from 1--64\,Hz and excluded 45--55\,Hz. This differs from the original range of frequencies above 0.5\,Hz up to 90\,Hz and was required to accommodate the 128-Hz recordings. Features from the three input signals were concatenated before LDA.

\paragraph{Model fitting and predictions}
The model was fitted to the WT training set using the authors' implementation. Hard predictions were obtained by Viterbi decoding over the complete recording, while probabilistic predictions were the forward--backward posterior marginals
$$
p(y_t=c\mid\mathbf{X}_{1:T}),
$$
where $\mathbf{X}_{1:T}$ denotes the complete sequence of LDA-projected observations. These probabilities incorporate both the Gaussian emissions and HMM transition dynamics; LDA class probabilities are not used directly for inference. MF1, Cohen's $\kappa$, and accuracy were computed from the Viterbi sequence, whereas LPD, Brier score, and ECE were computed from the forward--backward marginals. We use the same convention for Spindle below.

\begin{table}[t]
    \centering
    \caption{Overview of the baseline models and neural backbones.}
    \label{tab:baselines}
    \small
    \begin{tabular}{llll}
        \toprule
        Model
        & Architecture
        & Input
        & Context \\
        \midrule
        Somnotate
        & LDA+HMM
        & Spectrogram
        & Full recording \\
        Spindle
        & CNN+HMM
        & Spectrogram
        & Full recording \\
        U-Sleep
        & Fully convolutional
        & Raw signals
        & 35 windows \\
        CNN-GRU
        & CNN+GRU
        & Raw signals
        & 35 windows \\
        \bottomrule
    \end{tabular}
\end{table}

\subsection{Spindle}
\label{app:spindle}

We implemented Spindle following \citet{Miladinovic.etal-2019}. The model combines a CNN operating on spectrograms with a three-state HMM for temporal modeling. We did not implement the separate artifact detection CNN, as all predictive evaluations used the common three-class label space $\{\text{Wake},\text{NREM},\text{REM}\}$.

\paragraph{Spectrogram preprocessing}
Following \citet{Miladinovic.etal-2019}, signals were represented at 128\,Hz and transformed using 256-sample Hamming windows with a 16-sample step. The squared magnitude of the one-sided FFT gives a frequency resolution of $0.5$\,Hz and 32 time frames per 4-s scoring window. Spectrograms were computed continuously over each recording before division into scoring windows.

For the two ECoG/LFP channels, frequencies from $0.5$--$24$\,Hz were retained, yielding 48 coefficients. The published CNN architecture requires 24 frequency inputs, but the corresponding reduction is not specified by \citet{Miladinovic.etal-2019}; we therefore averaged adjacent coefficients pairwise in the linear-power domain. The resulting 24 bands were log-transformed and standardized within each recording. For EMG, spectral power from $0.5$--$30$\,Hz was summed, log-transformed, standardized over time, and repeated across the 24 frequency rows. The CNN input contained the target 4-s window and two neighboring windows on each side, giving an input size of $3\times24\times160$.

\paragraph{CNN training}
We reproduced the CNN architecture of \citet{Miladinovic.etal-2019}, including the two fully connected layers with dropout probability $0.5$. Training used the batch-dependent class-weighted cross-entropy described by the authors. Because the exact mapping from class ratio to weight is not specified, we used inverse-count weighting, $w_c=1/N_c$, with weights recomputed for each mini-batch. We used Adam with $\beta_1=0.9$, $\beta_2=0.99$, a mini-batch size of 100, and no weight decay.

The authors used a learning rate of $5\times10^{-5}$. We additionally evaluated $10^{-5}$ and $10^{-4}$ and found no meaningful differences in convergence or validation performance, and therefore retained the authors' setting. Following
\citet{Miladinovic.etal-2019}, training used early stopping after 20 consecutive validation evaluations without an improvement of at least $10^{-4}$, corresponding to a patience of 100 blocks.

\paragraph{HMM inference}
The CNN predictions were post-processed using a three-state HMM. Following \citet{Miladinovic.etal-2019}, we assumed equal state priors when converting the class-balanced CNN outputs to HMM emission scores. The CNN softmax probabilities could therefore be used directly, since division by a uniform state prior introduces only a common multiplicative factor. We likewise used the uniform initial-state distribution $\boldsymbol{\pi}=(1/3,1/3,1/3)$.

The original implementation assigns zero probability to some transitions, including Wake$\rightarrow$REM and REM$\rightarrow$NREM. Both occurred in the WT training annotations, so we instead estimated the full $3\times3$ transition matrix from the WT training labels.

As for Somnotate, hard predictions were obtained using Viterbi decoding and probabilistic predictions using forward--backward posterior marginals based on the same emissions, transition matrix, and initial-state probabilities. Accordingly, MF1, Cohen's $\kappa$, and accuracy were computed from the Viterbi sequence, whereas LPD, Brier score, and ECE were computed from the posterior marginals.

\subsection{Architecture selection for the neural network backbones}
\label{app:determin-backbones}

We next defined the two neural network backbones used for MAP estimation and the subsequent posterior approximation experiments. We used average validation MF1 as the model selection criterion.

\paragraph{Architecture-sweep training protocol}
All configurations in the architecture sweeps were trained with Adam using a learning rate of $10^{-4}$ and weight decay set to $10^{-5}$. Training was stopped when the validation cross-entropy had not improved by at least $10^{-4}$ for 100 consecutive blocks, and the checkpoint with the lowest validation cross-entropy was retained for comparison. This early-stopping protocol was used only for architecture selection.

\subsubsection{Adapting U-Sleep to rodent sleep scoring}
\label{app:usleep-selection}

\paragraph{Depth adaptation}
U-Sleep was originally developed for human sleep staging using 30-second epochs \citep{Perslev.etal-2021}, whereas sleep stages in the present study were scored in 4-second epochs. For a fixed number of context epochs and sampling frequency, the temporal input is therefore shorter by a factor of $30/4=7.5$.

The original U-Sleep encoder reduces temporal resolution through factor-two downsampling at each encoder level. Retaining all 12 encoder and decoder levels would therefore compress the shorter rodent input substantially more than in the original architecture. We reduced both the encoder and decoder from 12 to 9 blocks. Removing three factor-two downsampling stages reduces the total downsampling factor by $2^3=8$, approximately compensating for the $7.5$-fold reduction in epoch duration while otherwise retaining the original U-Sleep encoder--decoder design.

\paragraph{Model width}
We retained the width-scaling parameterization of U-Sleep \citep{Perslev.etal-2021}. Let $c_\ell$ denote the base number of filters at level $\ell$. The filter counts increase with encoder depth according to
$
c_{\ell+1}
=
\left\lfloor c_\ell\sqrt{2}\right\rfloor,
$
and are globally rescaled by a complexity factor $\alpha$,
$
\widetilde{c}_\ell(\alpha)
=
\left\lfloor c_\ell\sqrt{\alpha}\right\rfloor.
$
With the depth fixed at nine blocks and $c_0=8$, we evaluated
$$
\alpha\in\{0.25,0.5,1.0,2.0,3.0,4.0\}.
$$

As shown in Figure~\ref{fig:baseline-sizes}(a), validation MF1 varied less across the investigated widths than the variability across random initializations. The smallest configurations provided less fitting capacity, whereas increasing width substantially increased the number of parameters without providing a consistent validation improvement. We therefore selected $\alpha=1.0$ as a compact configuration within the high-performing region. The resulting model contains approximately $0.87$M parameters.

\subsubsection{CNN-GRU}
\label{app:cnn-gru-selection}

The second neural network backbone combines a convolutional encoder with a GRU for temporal modeling, following a common architecture pattern in automated sleep staging \citep{Yamabe.etal-2019,Olesen.etal-2021,Smith.etal-2025}.

\begin{figure}[t]
    \centering
    \includegraphics[width=\linewidth]{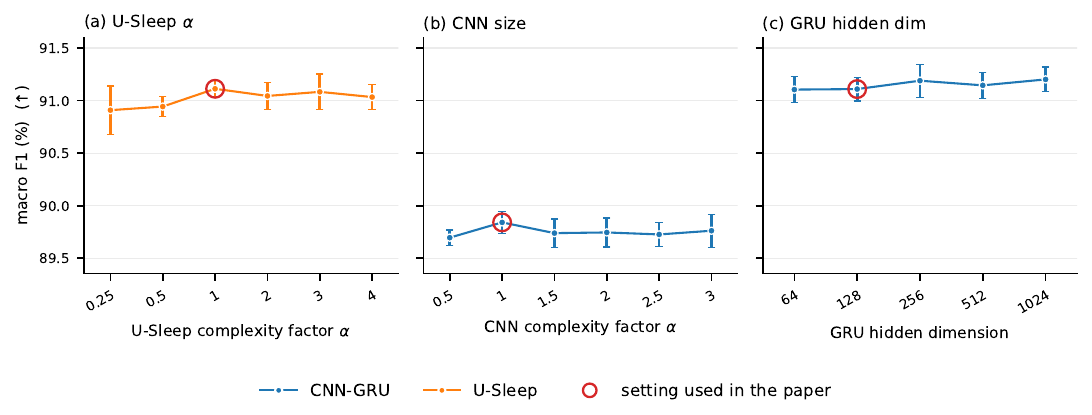}
    \caption{Validation MF1 as a function of model capacity during architecture selection.
    \textbf{(a)} U-Sleep complexity factor $\alpha$, evaluated across 10 random seeds.
    \textbf{(b)} CNN encoder complexity factor $\alpha$, evaluated across five random seeds.
    \textbf{(c)} GRU hidden dimension with the CNN encoder fixed at $\alpha=1.0$, evaluated across 10 random seeds.
    Points show means and error bars show one standard deviation across seeds.
    Validation performance varies only weakly once sufficient model capacity
    is reached. The selected configurations are $\alpha=1.0$ for U-Sleep,
    $\alpha=1.0$ for the CNN encoder, and $H=128$ for the GRU.
    }
    \label{fig:baseline-sizes}
\end{figure}

\paragraph{CNN encoder}
During preliminary development, we evaluated several one-dimensional convolutional encoder families, including residual architectures and the compact encoder proposed by \citet{Thapa.etal-2026}. We selected the latter because it provided competitive validation performance with a comparatively simple and parameter-efficient architecture. For this encoder, we introduced the same width scaling as used in the U-Sleep model and evaluated
$$
\alpha\in\{0.5,1.0,1.5,2.0,2.5,3.0\}.
$$
Validation performance was largely insensitive to increasing encoder width (Figure~\ref{fig:baseline-sizes}(b)), whereas parameter count increased substantially. We therefore fixed the encoder complexity at $\alpha=1.0$.

\paragraph{GRU hidden dimension}
With the encoder fixed at $\alpha=1.0$, we evaluated
$$
H\in\{64,128,256,512,1024\}
$$
for the GRU hidden dimension. Validation performance was again only weakly related to recurrent capacity (Figure~\ref{fig:baseline-sizes}(c)), while the total parameter count increased from approximately $1.12$M at $H=64$ to $8.39$M at $H=1024$. We selected $H=128$, yielding approximately $1.27$M parameters for the full CNN-GRU model. 

\subsection{MAP estimation and prior selection}
\label{app:map-prior-selection}

After fixing the two neural-network architectures, we selected a zero-mean isotropic Gaussian parameter prior separately for each backbone. The selected prior was subsequently held fixed across all posterior-approximation methods applied to that backbone.

\paragraph{MAP objective}
Let $\boldsymbol{\theta}_{\mathrm{wb}}$ denote the model weights and biases
assigned the prior
$$
p(\boldsymbol{\theta}_{\mathrm{wb}})
=
\mathcal{N}\!\left(
\boldsymbol{\theta}_{\mathrm{wb}}
\mid
\mathbf{0},
\sigma_p^2\mathbf{I}
\right).
$$
Normalization parameters were optimized but excluded from the prior. The potential energy $U(\boldsymbol{\theta})$ in Eq.~\eqref{eq:bayes} is defined using the full-data likelihood, whereas the implemented cross-entropy
$\mathcal{L}_{\mathrm{CE}}$ was averaged over the $N_{\mathrm{train}}$ training labels. Dividing the potential energy by $N_{\mathrm{train}}$ does not change its minimum, giving the MAP objective
\begin{equation}
\widehat{\boldsymbol{\theta}}_{\mathrm{MAP}}
=
\argmin_{\boldsymbol{\theta}}
\left[
\mathcal{L}_{\mathrm{CE}}(\boldsymbol{\theta})
+
\frac{\lambda}{2}
\left\lVert
\boldsymbol{\theta}_{\mathrm{wb}}
\right\rVert_2^2
\right],
\qquad
\lambda
=
\frac{1}{N_{\mathrm{train}}\sigma_p^2}.
\label{eq:map-training-objective}
\end{equation}
Thus, $\lambda$ is the mean-loss $L_2$ coefficient corresponding to the Gaussian prior variance $\sigma_p^2$.

\paragraph{Prior selection}
For each backbone, we evaluated
\[
\lambda
\in
\left\{
3\times10^{-5},
10^{-4},
3\times10^{-4},
10^{-3},
3\times10^{-3}
\right\},
\]
and selected the value with the highest mean validation MF1 (Figure~\ref{fig:bayes-sensitivity}(a)). This yielded $\lambda=10^{-3}$ for CNN-GRU and $\lambda=3\times10^{-3}$ for U-Sleep, corresponding to prior variances $\sigma_p^2=2.83\times10^{-4}$ and $\sigma_p^2=9.42\times10^{-5}$, respectively, using $\sigma_p^2=(N_{\mathrm{train}}\lambda)^{-1}$ with $N_{\mathrm{train}}=3{,}537{,}813$. All configurations were trained without early stopping for approximately 57 passes through the training set, and the final checkpoint was retained as the MAP estimate $\widehat{\boldsymbol{\theta}}_{\mathrm{MAP}}$. The final MAP models are summarized in Table~\ref{tab:final-map-models}.

\begin{table}[t]
    \centering
    \caption{Final MAP model hyperparameters.}
    \label{tab:final-map-models}
    \small
    \begin{tabular}{lllll}
        \toprule
        Model
        & Architecture
        & Parameters
        & $\lambda$
        & $\sigma_p^2$ \\
        \midrule
        U-Sleep
        & 9 blocks, $\alpha=1.0$
        & $\sim0.87$M
        & $3\times10^{-3}$ 
        & $9.42\times10^{-5}$\\
        CNN-GRU
        & Encoder $\alpha=1.0$, GRU $H=128$
        & $\sim1.27$M
        & $10^{-3}$ 
        & $2.83\times10^{-4}$ \\
        \bottomrule
    \end{tabular}
\end{table}

\newpage
\section{Implementation and Tuning of Approximate Inference Methods}
\label{app:bayes}
\begin{figure}[p]
    \centering
    \includegraphics[
        width=\linewidth,
        height=0.90\textheight,
        keepaspectratio
    ]{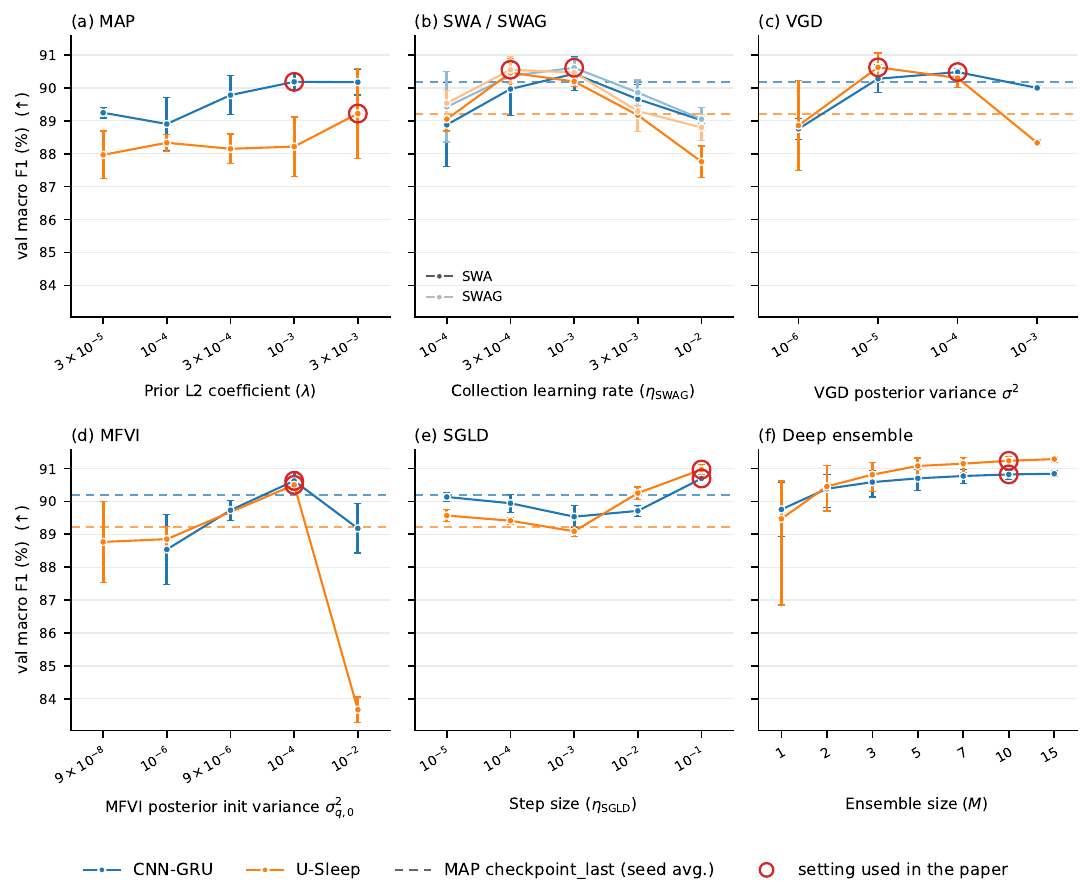}
    \caption{
    Sensitivity of validation MF1 to the primary method-specific hyperparameters. For sampling-based posterior approximations, predictive probabilities were estimated using $S=30$ samples.
    \textbf{(a)} MAP prior strength $\lambda$ (equivalently, prior variance $\sigma_p^2$).
    \textbf{(b)} Shared SWA/SWAG collection learning rate.
    \textbf{(c)} VGD posterior variance.
    \textbf{(d)} MFVI initial posterior variance $\sigma_{q,0}^2$.
    \textbf{(e)} SGLD batch-scaled step size $\eta_{\mathrm{batch}}$. 
    \textbf{(f)} DE size. Error bars show one standard deviation across seeds in \textbf{(a--e)} and the central $95\%$ empirical interval across 100 random ensemble subsets in \textbf{(f)}.
    }
    \label{fig:bayes-sensitivity}
\end{figure}

This section describes the implementation and tuning of the posterior approximation and predictive-averaging methods. Figure~\ref{fig:bayes-sensitivity} summarizes the primary sensitivity analyses, and Table~\ref{tab:bayes-hyperparameters} reports the selected hyperparameters, optimization budgets, and predictive-integration settings.

\begin{table}[t]
\centering
\caption{
Selected hyperparameters, optimization budgets, and predictive-integration
settings for the controlled inference comparison. All tunable settings were
selected using mean validation MF1.
}
\label{tab:bayes-hyperparameters}
\scriptsize
\setlength{\tabcolsep}{3pt}
\renewcommand{\arraystretch}{1.18}
\begin{tabularx}{\linewidth}{
    @{}
    >{\raggedright\arraybackslash}p{0.12\linewidth}
    >{\centering\arraybackslash}p{0.14\linewidth}
    >{\centering\arraybackslash}p{0.14\linewidth}
    >{\centering\arraybackslash}p{0.15\linewidth}
    >{\raggedright\arraybackslash}X
    @{}
}
\toprule
Method
& CNN-GRU
& U-Sleep
& Training/collection budget
& Shared settings \\
\midrule

MAP
& $\sigma_p^2 = 2.83\times10^{-4}$
& $\sigma_p^2 = 9.42\times10^{-5}$
& 2000 blocks
& Final checkpoint; deterministic prediction \\

\addlinespace[3pt]
SWA / SWAG
& $\eta_{\mathrm{collect}}=10^{-3}$
& $\eta_{\mathrm{collect}}=3\times10^{-4}$
& 5 transition $+$ 200 collection blocks
& Same trajectory for SWA and SWAG; one iterate per collection block; $R=100$; SWAG $S=30$; BN statistics recomputed using $10\%$ of the training data \\

\addlinespace[3pt]
LLLA
& Full covariance
& Full covariance
& --
& MAP prior variance; GLM predictive with $S=30$ Monte Carlo samples \\

\addlinespace[3pt]
VGD
& $\sigma^2=10^{-4}$
& $\sigma^2=10^{-5}$
& 3000 blocks
& Random initialization; MAP prior retained; $S_{\mathrm{train}}=1$, $S_{\mathrm{val}}=10$, final $S=30$; learning rate $=10^{-4}$ \\

\addlinespace[3pt]
MFVI
& $\sigma_{q,0}^2=10^{-4}$
& $\sigma_{q,0}^2=10^{-4}$
& 3000 blocks
& MAP initialization of posterior means; MAP prior retained; $S_{\mathrm{train}}=1$, $S_{\mathrm{val}}=10$, final $S=30$; learning rate $=10^{-4}$ \\

\addlinespace[3pt]
SGLD
& $\eta_{\mathrm{batch}}=10^{-1}$
& $\eta_{\mathrm{batch}}=10^{-1}$
& 3000 blocks
& MAP initialization and prior; burn-in 1000 blocks; states stored every 10 blocks; $S=30$ selected uniformly from 200 post-burn-in states \\

\addlinespace[3pt]
Deep ensemble
& $M=10$
& $M=10$
& 2000 blocks per member
& Pool of 20 independently trained MAP models; 100 subset resamples during size selection \\

\addlinespace[3pt]
MultiSWA / MultiSWAG
& $M=10$
& $M=10$
& 5 transition $+$ 200 collection blocks per component
& Same settings as single-model SWA/SWAG; $R=100$; MultiSWAG $S=30$ samples per component; BN-statistics recomputation \\

\bottomrule
\end{tabularx}
\end{table}

\subsection{SWA and SWAG}
\label{app:swag}

\paragraph{Initialization and collection}
For each backbone and random seed, we initialized the SWA and SWAG collection trajectory from the corresponding final MAP checkpoint described in Section~\ref{app:map-prior-selection}. We first ramped the learning rate from the MAP training learning rate to the target collection learning rate over five transition blocks, during which no parameter iterates were retained.

We then continued optimization for 200 collection blocks at the collection learning rate and stored one parameter iterate after each block. SWA was constructed by averaging all 200 iterates. For SWAG, the diagonal covariance component was estimated from all iterates, while the final $R=100$ iterates were used for the low-rank component.

Because both backbones contain batch-normalization layers, we recomputed the running statistics using $10\%$ of the training data after changing the parameter values. This was performed once for the SWA solution and separately for each sampled SWAG model. Final SWAG predictions were obtained by averaging $S=30$ parameter samples.

\paragraph{Collection learning rate selection}
The collection learning rate controls the extent to which the continued SGD trajectory explores the local parameter region around the MAP solution. We evaluated
\[
\eta_{\mathrm{collect}}
\in
\left\{
10^{-4},
3\times10^{-4},
10^{-3},
3\times10^{-3},
10^{-2}
\right\}.
\]
For each candidate learning rate, SWA and SWAG were constructed from the same trajectory. We therefore selected one collection learning rate per backbone and used it for both methods, enabling a paired comparison of the SWA point estimate and the corresponding SWAG approximation. Based on the validation sensitivity analysis in Figure~\ref{fig:bayes-sensitivity}(b), we selected $ \eta_{\mathrm{collect}} = 10^{-3}$ for CNN-GRU and $\eta_{\mathrm{collect}} = 3\times10^{-4}$ for U-Sleep.

\subsection{Last-layer Laplace approximation}
\label{app:laplace}

We fitted the LLLA defined in Section~\ref{sec:inference} to the final MAP checkpoints of both backbones, CNN-GRU and U-Sleep, using the PyTorch implementation of \citet{Daxberger.etal-2021}. All other parameters were fixed at their MAP values, and a full-covariance Gaussian approximation was fitted to the weights and biases of the final classification layer. We retained the Gaussian prior selected for each MAP model.

Predictive probabilities were computed from the linearized generalized linear model predictive distribution using Monte Carlo sampling. Specifically, we drew $S=30$ samples from the Gaussian predictive distribution over the logits, applied the softmax to each sample, and averaged the resulting class probabilities. The same samples were used for the decomposition of predictive uncertainty into aleatoric and epistemic components.

The implementation requires the final layer to be linear, which holds for CNN-GRU. The final layer of U-Sleep is a $1\times1$ convolution, which applies the same linear map to the features of every output window. For the Laplace fit, we therefore represented it as an equivalent linear layer with the same weights and biases. This changes neither the trained parameters nor the U-Sleep predictions, which we verified numerically.

\subsection{Variational gradient descent}
\label{app:vgd}

VGD used the fixed-variance Gaussian approximation and objective defined in Section~\ref{sec:inference}. Gaussian perturbations were applied to model weights and biases, while normalization parameters remained deterministic. For each backbone, VGD retained the Gaussian prior selected for the corresponding MAP model but was optimized independently from a random initialization rather than initialized from the MAP solution.

We optimized the variational mean for 3000 blocks using Adam with a learning rate of $10^{-4}$. The expected objective gradient was estimated using $S_{\mathrm{train}}=1$ Gaussian perturbations per optimization step. Validation performance was estimated using $S_{\mathrm{val}}=10$ posterior samples, and final posterior-predictive probabilities were computed using the common sampling budget of $S=30$.

\paragraph{Posterior-variance selection}
We treated the fixed posterior variance $\sigma^2$ as the principal VGD-specific hyperparameter, although \citet{Ghosh.etal-2025} show that both the learning rate, the number of Monte Carlo samples and the posterior variance control the sharpness of the solution. We evaluated $\sigma^2\in\{0,10^{-6},10^{-5},10^{-4},10^{-3}\}$, where $\sigma^2=0$ served as a deterministic reference. Based on mean validation MF1, we selected $\sigma^2=10^{-4}$ for CNN-GRU and $\sigma^2=10^{-5}$ for U-Sleep (Figure~\ref{fig:bayes-sensitivity}(c)).

\subsection{Mean-field variational inference}
\label{app:mfvi}
MFVI used the fully factorized Gaussian approximation and variational objective defined in Section~\ref{sec:inference}. Variational distributions were assigned to model weights and biases, while normalization parameters remained deterministic. Posterior standard deviations were parameterized as $\sigma_i=\operatorname{softplus}(\rho_i)$.

For each backbone, we retained the zero-mean isotropic Gaussian prior selected for the corresponding MAP model. Because the likelihood term was implemented as mean cross-entropy, the KL divergence was scaled by $1/N_{\mathrm{train}}$. No additional $L_2$ penalty was applied, since regularization toward the Gaussian prior was already represented by the KL term.

The posterior means were initialized from the corresponding final MAP checkpoint, $\boldsymbol{\mu}_0=\widehat{\boldsymbol{\theta}}_{\mathrm{MAP}}$, and all posterior variances were initialized to a common value $\sigma_{q,0}^2$. This initial variance was the only MFVI-specific hyperparameter tuned. For U-Sleep, we evaluated $\sigma_{q,0}^2\in\{9\times10^{-8},10^{-6},10^{-4},10^{-2}\}$; for CNN-GRU, we evaluated $\sigma_{q,0}^2\in\{10^{-6},9\times10^{-6},10^{-4},10^{-2}\}$. Based on mean validation MF1, we selected $\sigma_{q,0}^2=10^{-4}$ for both U-Sleep and CNN-GRU (Figure~\ref{fig:bayes-sensitivity}(d)).

We optimized the variational parameters for 3000 blocks using Adam with a learning rate of $10^{-4}$ and one Monte Carlo sample per optimization step. Validation performance was estimated using $S_{\mathrm{val}}=10$ samples, and final posterior-predictive probabilities were computed using $S=30$ samples.

\subsection{Stochastic gradient Langevin dynamics}
\label{app:sgld}

We implemented SGLD using the update in Eq.~\eqref{eq:sgld}. For each backbone and random seed, the chain was initialized from the corresponding final MAP checkpoint and used the same backbone-specific Gaussian prior.

Because SGLD is defined using the full-data potential energy, its mini-batch estimate was
\begin{equation}
\widehat{U}_{\mathcal{B}}(\boldsymbol{\theta})
=
N_{\mathrm{train}}\,
\overline{\ell}_{\mathcal{B}}(\boldsymbol{\theta})
+
\frac{1}{2\sigma_p^2}
\left\lVert
\boldsymbol{\theta}_{\mathrm{wb}}
\right\rVert_2^2,
\label{eq:sgld-energy}
\end{equation}
where $\overline{\ell}_{\mathcal{B}}$ is the mean cross-entropy over mini-batch $\mathcal{B}$. The corresponding prior precision is $\sigma_p^{-2}=N_{\mathrm{train}}\lambda$. Normalization parameters remained deterministic and were excluded from the prior.

\begin{figure}[t]
    \centering
    \includegraphics[width=\linewidth]
    {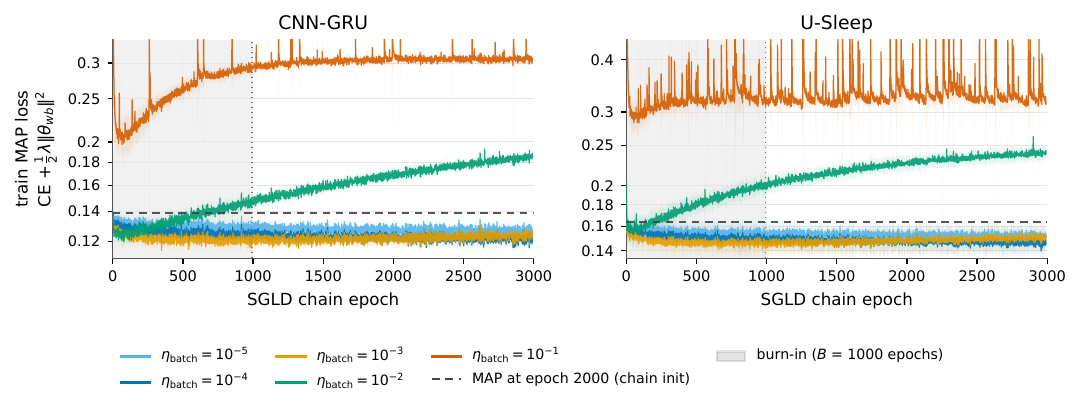}
    \caption{
    Mean-scale MAP-objective trajectories for SGLD chains initialized
    from the final MAP checkpoints. Colors indicate candidate values of
    $\eta_{\mathrm{batch}}$, dashed horizontal lines show the objective
    values at initialization, and the shaded region denotes the
    1000-block burn-in period. The selected value
    $\eta_{\mathrm{batch}}=10^{-1}$ was chosen using validation
    MF1; the trajectories were used only to identify and remove the
    clearest initialization transient, not to claim formal convergence.
    }
    \label{fig:sgld-diagnostics}
\end{figure}

\paragraph{Step-size selection}
We used a constant step size as a practical finite-chain approximation \citep{Welling.Teh-2011,Izmailov.etal-2021}. To express the step size on the same mean-loss scale used during MAP training, we set
$$
\eta
=
\frac{\eta_{\mathrm{batch}}}{N_{\mathrm{train}}},
$$
following \citet{Chen.etal-2014}. We evaluated
$$
\eta_{\mathrm{batch}}
\in
\left\{
10^{-5},
10^{-4},
10^{-3},
10^{-2},
10^{-1}
\right\}.
$$
Based on mean validation MF1, we selected $\eta_{\mathrm{batch}}=10^{-1}$ for both CNN-GRU and U-Sleep (Figure~\ref{fig:bayes-sensitivity}(e)). The objective trajectories in Figure~\ref{fig:sgld-diagnostics} were used as sampling diagnostics rather than for step-size selection.

\paragraph{Burn-in and posterior prediction}
Each chain was run for 3000 blocks, with one parameter state retained every 10 blocks. At the selected step size, the chains exhibited an initial transition away from the MAP solution, followed by a higher-objective regime with less systematic drift, although stochastic fluctuations and occasional spikes remained. We therefore discarded the first 1000 chain blocks, corresponding to the first 100 retained states, as burn-in.

This left 200 post-burn-in states, from which we selected $S=30$ uniformly spaced samples, resulting in approximately one selected sample every 67 blocks. Batch-normalization running statistics were stored with each retained state and used directly during prediction. Because we used a constant step size and a finite chain, the retained states are treated as practical approximate posterior samples. The burn-in period removes the clearest initialization transient but is not intended as evidence of formal Markov-chain convergence.

\subsection{Deep ensembles}
\label{app:ensembles}
For each backbone, we trained a pool of $20$ MAP models from independent random initializations. All members used the same architecture, training procedure, and backbone-specific Gaussian prior, and the final checkpoint from each run was retained. Ensemble predictions were obtained by equally averaging the member-wise predictive probabilities.

\paragraph{Ensemble-size selection}
We evaluated ensemble sizes $M\in\{1,2,3,5,7,10,15\}$. For each value of $M$, we drew $100$ random $M$-member ensembles from the pool of $20$ MAP models and evaluated their validation predictions. Mean validation MF1 improved with ensemble size, with diminishing gains beyond approximately $M=10$ (Figure~\ref{fig:bayes-sensitivity}(f)). We therefore used $M=10$ for the final DE experiments.

\subsection{MultiSWA and MultiSWAG}
\label{app:multi-swag}
We constructed MultiSWA and MultiSWAG from the same $M=10$ independently trained MAP models used for the corresponding DE. Starting from each MAP checkpoint, we applied the same optimization trajectory, collection learning rate, collection length, low-rank dimension, and batch-normalization recalibration procedure as for the corresponding single-model SWA and SWAG methods described in Section~\ref{app:swag}. No additional hyperparameters were tuned for either multi-basin method.

MultiSWA equally averaged predictions from the $M$ resulting SWA point estimates. For MultiSWAG, we drew $S=30$ parameter samples from each SWAG component, matching the sampling budget used for single-model SWAG. Predictions were first averaged over samples within each component and then averaged equally across the $M$ components. With $M=10$, this corresponded to $300$ parameter samples in total.

\section{ADDITIONAL RESULTS}
\label{app:additional-results}

%%%%%%%% FULL ID + OOD performance
\begin{table*}[t]
\centering
\caption{\textbf{In-distribution performance on the WT test set.} Values are mean $\pm$ 95\% CIs across seeds; for DE, MultiSWA, and MultiSWAG, $\pm$ denotes half the empirical 2.5--97.5\% interval across 100 resampled 10-member ensembles. Accuracy, MF1, $\kappa$, Brier score, and ECE are reported in percent; LPD and entropy are in nats. \textbf{Bold} and \underline{underlining} denote the best overall and within each model group, respectively; entropy is descriptive and not ranked.}
\label{tab:id-results-wide}
\scriptsize
\setlength{\tabcolsep}{4pt}
\begin{tabular}{lccccccc}
\toprule
Model & Acc.\,$\uparrow$ & MF1\,$\uparrow$ & $\kappa$\,$\uparrow$ & LPD\,$\uparrow$ & Brier\,$\downarrow$ & ECE\,$\downarrow$ & Entropy \\
\midrule
\multicolumn{8}{@{}l}{\itshape U-Sleep} \\
MAP & $93.8\,{\scriptstyle\pm1.7}$ & $89.7\,{\scriptstyle\pm1.5}$ & $88.6\,{\scriptstyle\pm3.1}$ & ${\scriptstyle-}.186\,{\scriptstyle\pm.048}$ & $9.4\,{\scriptstyle\pm2.6}$ & $1.7\,{\scriptstyle\pm0.7}$ & $.168\,{\scriptstyle\pm.041}$ \\
Temp. & $93.8\,{\scriptstyle\pm1.7}$ & $89.7\,{\scriptstyle\pm1.5}$ & $88.6\,{\scriptstyle\pm3.1}$ & ${\scriptstyle-}.185\,{\scriptstyle\pm.046}$ & $9.4\,{\scriptstyle\pm2.5}$ & $1.6\,{\scriptstyle\pm0.4}$ & $.189\,{\scriptstyle\pm.025}$ \\
SWA & $95.3\,{\scriptstyle\pm0.4}$ & $91.5\,{\scriptstyle\pm0.3}$ & $91.4\,{\scriptstyle\pm0.7}$ & ${\scriptstyle-}.141\,{\scriptstyle\pm.009}$ & $7.1\,{\scriptstyle\pm0.5}$ & $\textbf{\underline{1.4}}\,{\scriptstyle\pm0.1}$ & $.149\,{\scriptstyle\pm.010}$ \\
SWAG & $95.4\,{\scriptstyle\pm0.3}$ & $91.7\,{\scriptstyle\pm0.2}$ & $91.6\,{\scriptstyle\pm0.5}$ & ${\scriptstyle-}.138\,{\scriptstyle\pm.005}$ & $6.9\,{\scriptstyle\pm0.3}$ & $1.6\,{\scriptstyle\pm0.1}$ & $.162\,{\scriptstyle\pm.009}$ \\
Laplace & $93.8\,{\scriptstyle\pm1.7}$ & $89.7\,{\scriptstyle\pm1.5}$ & $88.6\,{\scriptstyle\pm3.1}$ & ${\scriptstyle-}.186\,{\scriptstyle\pm.048}$ & $9.4\,{\scriptstyle\pm2.6}$ & $1.7\,{\scriptstyle\pm0.7}$ & $.168\,{\scriptstyle\pm.041}$ \\
VGD & $94.2\,{\scriptstyle\pm1.8}$ & $90.6\,{\scriptstyle\pm1.4}$ & $89.4\,{\scriptstyle\pm3.4}$ & ${\scriptstyle-}.181\,{\scriptstyle\pm.044}$ & $9.1\,{\scriptstyle\pm2.6}$ & $3.7\,{\scriptstyle\pm1.0}$ & $.254\,{\scriptstyle\pm.041}$ \\
MFVI & $94.4\,{\scriptstyle\pm0.8}$ & $90.5\,{\scriptstyle\pm0.6}$ & $89.7\,{\scriptstyle\pm1.5}$ & ${\scriptstyle-}.213\,{\scriptstyle\pm.019}$ & $9.9\,{\scriptstyle\pm1.2}$ & $7.3\,{\scriptstyle\pm1.5}$ & $.366\,{\scriptstyle\pm.045}$ \\
SGLD & $95.4\,{\scriptstyle\pm0.0}$ & $91.7\,{\scriptstyle\pm0.0}$ & $91.6\,{\scriptstyle\pm0.1}$ & ${\scriptstyle-}.152\,{\scriptstyle\pm.003}$ & $7.2\,{\scriptstyle\pm0.1}$ & $3.0\,{\scriptstyle\pm0.3}$ & $.231\,{\scriptstyle\pm.012}$ \\
DE & $95.4\,{\scriptstyle\pm0.2}$ & $91.7\,{\scriptstyle\pm0.2}$ & $91.6\,{\scriptstyle\pm0.4}$ & ${\scriptstyle-}.142\,{\scriptstyle\pm.006}$ & $7.0\,{\scriptstyle\pm0.3}$ & $2.1\,{\scriptstyle\pm0.3}$ & $.178\,{\scriptstyle\pm.013}$ \\
MultiSWA & $95.6\,{\scriptstyle\pm0.1}$ & $91.9\,{\scriptstyle\pm0.0}$ & $91.9\,{\scriptstyle\pm0.1}$ & ${\scriptstyle-}.133\,{\scriptstyle\pm.002}$ & $6.7\,{\scriptstyle\pm0.1}$ & $1.5\,{\scriptstyle\pm0.1}$ & $.151\,{\scriptstyle\pm.005}$ \\
MultiSWAG & $\textbf{\underline{95.6}}\,{\scriptstyle\pm0.0}$ & $\textbf{\underline{91.9}}\,{\scriptstyle\pm0.0}$ & $\textbf{\underline{92.0}}\,{\scriptstyle\pm0.1}$ & ${\scriptstyle-}\textbf{\underline{.133}}\,{\scriptstyle\pm.002}$ & $\textbf{\underline{6.6}}\,{\scriptstyle\pm0.1}$ & $1.7\,{\scriptstyle\pm0.1}$ & $.164\,{\scriptstyle\pm.004}$ \\
\addlinespace[2pt]
\multicolumn{8}{@{}l}{\itshape CNN-GRU} \\
MAP & $94.6\,{\scriptstyle\pm0.2}$ & $90.5\,{\scriptstyle\pm0.2}$ & $90.1\,{\scriptstyle\pm0.4}$ & ${\scriptstyle-}.161\,{\scriptstyle\pm.006}$ & $8.1\,{\scriptstyle\pm0.3}$ & $1.5\,{\scriptstyle\pm0.2}$ & $.123\,{\scriptstyle\pm.013}$ \\
Temp. & $94.6\,{\scriptstyle\pm0.2}$ & $90.5\,{\scriptstyle\pm0.2}$ & $90.1\,{\scriptstyle\pm0.4}$ & ${\scriptstyle-}.157\,{\scriptstyle\pm.003}$ & $8.1\,{\scriptstyle\pm0.2}$ & $1.5\,{\scriptstyle\pm0.1}$ & $.168\,{\scriptstyle\pm.007}$ \\
SWA & $95.4\,{\scriptstyle\pm0.2}$ & $91.7\,{\scriptstyle\pm0.2}$ & $91.6\,{\scriptstyle\pm0.3}$ & ${\scriptstyle-}.135\,{\scriptstyle\pm.007}$ & $6.9\,{\scriptstyle\pm0.3}$ & $1.6\,{\scriptstyle\pm0.3}$ & $.153\,{\scriptstyle\pm.019}$ \\
SWAG & $95.5\,{\scriptstyle\pm0.2}$ & $91.8\,{\scriptstyle\pm0.2}$ & $91.7\,{\scriptstyle\pm0.3}$ & ${\scriptstyle-}.137\,{\scriptstyle\pm.008}$ & $6.9\,{\scriptstyle\pm0.4}$ & $2.0\,{\scriptstyle\pm0.4}$ & $.171\,{\scriptstyle\pm.021}$ \\
Laplace & $94.6\,{\scriptstyle\pm0.2}$ & $90.5\,{\scriptstyle\pm0.2}$ & $90.1\,{\scriptstyle\pm0.4}$ & ${\scriptstyle-}.161\,{\scriptstyle\pm.006}$ & $8.1\,{\scriptstyle\pm0.3}$ & $1.5\,{\scriptstyle\pm0.2}$ & $.123\,{\scriptstyle\pm.013}$ \\
VGD & $95.2\,{\scriptstyle\pm0.3}$ & $91.5\,{\scriptstyle\pm0.3}$ & $91.3\,{\scriptstyle\pm0.5}$ & ${\scriptstyle-}.148\,{\scriptstyle\pm.007}$ & $7.4\,{\scriptstyle\pm0.3}$ & $2.6\,{\scriptstyle\pm0.3}$ & $.196\,{\scriptstyle\pm.009}$ \\
MFVI & $94.7\,{\scriptstyle\pm0.3}$ & $90.8\,{\scriptstyle\pm0.2}$ & $90.3\,{\scriptstyle\pm0.5}$ & ${\scriptstyle-}.172\,{\scriptstyle\pm.009}$ & $8.5\,{\scriptstyle\pm0.5}$ & $3.8\,{\scriptstyle\pm0.5}$ & $.244\,{\scriptstyle\pm.019}$ \\
SGLD & $95.1\,{\scriptstyle\pm0.2}$ & $91.2\,{\scriptstyle\pm0.1}$ & $91.0\,{\scriptstyle\pm0.3}$ & ${\scriptstyle-}.143\,{\scriptstyle\pm.005}$ & $7.4\,{\scriptstyle\pm0.3}$ & $1.6\,{\scriptstyle\pm0.1}$ & $.151\,{\scriptstyle\pm.008}$ \\
DE & $95.1\,{\scriptstyle\pm0.1}$ & $91.3\,{\scriptstyle\pm0.1}$ & $91.1\,{\scriptstyle\pm0.3}$ & ${\scriptstyle-}.141\,{\scriptstyle\pm.004}$ & $7.3\,{\scriptstyle\pm0.2}$ & $\underline{1.4}\,{\scriptstyle\pm0.1}$ & $.135\,{\scriptstyle\pm.005}$ \\
MultiSWA & $95.5\,{\scriptstyle\pm0.0}$ & $91.9\,{\scriptstyle\pm0.0}$ & $91.8\,{\scriptstyle\pm0.1}$ & ${\scriptstyle-}\underline{.133}\,{\scriptstyle\pm.001}$ & $\underline{6.8}\,{\scriptstyle\pm0.1}$ & $1.6\,{\scriptstyle\pm0.1}$ & $.153\,{\scriptstyle\pm.004}$ \\
MultiSWAG & $\underline{95.5}\,{\scriptstyle\pm0.0}$ & $\underline{91.9}\,{\scriptstyle\pm0.0}$ & $\underline{91.8}\,{\scriptstyle\pm0.1}$ & ${\scriptstyle-}.136\,{\scriptstyle\pm.001}$ & $6.8\,{\scriptstyle\pm0.1}$ & $2.0\,{\scriptstyle\pm0.1}$ & $.172\,{\scriptstyle\pm.004}$ \\
\addlinespace[2pt]
\multicolumn{8}{@{}l}{\itshape Other baselines} \\
Somnotate & $\underline{95.1}$ & $\underline{90.4}$ & $\underline{91.1}$ & ${\scriptstyle-}\underline{.248}$ & $\underline{8.4}$ & $\underline{3.7}$ & $.032$ \\
Spindle & $94.2\,{\scriptstyle\pm0.1}$ & $89.2\,{\scriptstyle\pm0.4}$ & $89.5\,{\scriptstyle\pm0.2}$ & ${\scriptstyle-}.301\,{\scriptstyle\pm.005}$ & $10.1\,{\scriptstyle\pm0.2}$ & $4.4\,{\scriptstyle\pm0.1}$ & $.037\,{\scriptstyle\pm.003}$ \\
\bottomrule
\end{tabular}
\end{table*}            % \ref{tab:id-results-wide}
\begin{table*}[t]
\centering
\caption{\textbf{Out-of-distribution performance.} Metrics are averaged over the five OOD cohorts within each training seed. Values are mean $\pm$ 95\% CIs across seeds; for DE, MultiSWA, and MultiSWAG, $\pm$ denotes half the empirical 2.5--97.5\% interval across 100 resampled 10-member ensembles. Accuracy, MF1, $\kappa$, Brier score, and ECE are reported in percent; LPD and entropy are in nats. \textbf{Bold} and \underline{underlining} denote the best overall and within each model group, respectively; entropy is descriptive and not ranked.}
\label{tab:ood-results-wide}
\scriptsize
\setlength{\tabcolsep}{4pt}
\begin{tabular}{lccccccc}
\toprule
Model & Acc.\,$\uparrow$ & MF1\,$\uparrow$ & $\kappa$\,$\uparrow$ & LPD\,$\uparrow$ & Brier\,$\downarrow$ & ECE\,$\downarrow$ & Entropy \\
\midrule
\multicolumn{8}{@{}l}{\itshape U-Sleep} \\
MAP & $84.6\,{\scriptstyle\pm3.5}$ & $76.0\,{\scriptstyle\pm4.4}$ & $72.1\,{\scriptstyle\pm6.3}$ & ${\scriptstyle-}.481\,{\scriptstyle\pm.149}$ & $24.2\,{\scriptstyle\pm6.7}$ & $7.7\,{\scriptstyle\pm4.5}$ & $.224\,{\scriptstyle\pm.040}$ \\
Temp. & $84.6\,{\scriptstyle\pm3.5}$ & $76.0\,{\scriptstyle\pm4.4}$ & $72.1\,{\scriptstyle\pm6.3}$ & ${\scriptstyle-}.460\,{\scriptstyle\pm.122}$ & $23.9\,{\scriptstyle\pm6.3}$ & $7.2\,{\scriptstyle\pm3.9}$ & $.246\,{\scriptstyle\pm.015}$ \\
SWA & $85.8\,{\scriptstyle\pm3.1}$ & $78.0\,{\scriptstyle\pm3.2}$ & $74.3\,{\scriptstyle\pm5.9}$ & ${\scriptstyle-}.435\,{\scriptstyle\pm.105}$ & $21.7\,{\scriptstyle\pm4.8}$ & $6.3\,{\scriptstyle\pm2.6}$ & $.243\,{\scriptstyle\pm.031}$ \\
SWAG & $86.7\,{\scriptstyle\pm2.7}$ & $79.2\,{\scriptstyle\pm2.8}$ & $75.7\,{\scriptstyle\pm5.3}$ & ${\scriptstyle-}.376\,{\scriptstyle\pm.060}$ & $19.8\,{\scriptstyle\pm3.6}$ & $5.5\,{\scriptstyle\pm1.9}$ & $.282\,{\scriptstyle\pm.027}$ \\
Laplace & $84.6\,{\scriptstyle\pm3.5}$ & $76.0\,{\scriptstyle\pm4.4}$ & $72.1\,{\scriptstyle\pm6.3}$ & ${\scriptstyle-}.481\,{\scriptstyle\pm.149}$ & $24.2\,{\scriptstyle\pm6.7}$ & $7.7\,{\scriptstyle\pm4.5}$ & $.224\,{\scriptstyle\pm.040}$ \\
VGD & $82.1\,{\scriptstyle\pm7.5}$ & $75.2\,{\scriptstyle\pm6.2}$ & $68.2\,{\scriptstyle\pm12.0}$ & ${\scriptstyle-}.473\,{\scriptstyle\pm.202}$ & $25.7\,{\scriptstyle\pm10.6}$ & $8.1\,{\scriptstyle\pm3.9}$ & $.393\,{\scriptstyle\pm.038}$ \\
MFVI & $85.1\,{\scriptstyle\pm1.9}$ & $79.6\,{\scriptstyle\pm2.0}$ & $73.2\,{\scriptstyle\pm3.5}$ & ${\scriptstyle-}.415\,{\scriptstyle\pm.038}$ & $22.6\,{\scriptstyle\pm2.5}$ & $10.0\,{\scriptstyle\pm1.9}$ & $.497\,{\scriptstyle\pm.054}$ \\
SGLD & $86.8\,{\scriptstyle\pm1.1}$ & $80.6\,{\scriptstyle\pm1.3}$ & $76.1\,{\scriptstyle\pm2.0}$ & ${\scriptstyle-}.362\,{\scriptstyle\pm.028}$ & $19.4\,{\scriptstyle\pm1.5}$ & $6.2\,{\scriptstyle\pm0.7}$ & $.361\,{\scriptstyle\pm.007}$ \\
DE & $\textbf{\underline{89.0}}\,{\scriptstyle\pm1.2}$ & $82.1\,{\scriptstyle\pm1.5}$ & $\underline{79.5}\,{\scriptstyle\pm2.4}$ & ${\scriptstyle-}\textbf{\underline{.299}}\,{\scriptstyle\pm.019}$ & $\textbf{\underline{16.2}}\,{\scriptstyle\pm1.3}$ & $4.7\,{\scriptstyle\pm0.8}$ & $.319\,{\scriptstyle\pm.025}$ \\
MultiSWA & $88.8\,{\scriptstyle\pm0.9}$ & $82.2\,{\scriptstyle\pm0.9}$ & $79.3\,{\scriptstyle\pm2.0}$ & ${\scriptstyle-}.315\,{\scriptstyle\pm.018}$ & $16.6\,{\scriptstyle\pm1.2}$ & $\textbf{\underline{4.4}}\,{\scriptstyle\pm0.8}$ & $.286\,{\scriptstyle\pm.012}$ \\
MultiSWAG & $88.9\,{\scriptstyle\pm0.8}$ & $\textbf{\underline{82.5}}\,{\scriptstyle\pm0.8}$ & $79.4\,{\scriptstyle\pm1.8}$ & ${\scriptstyle-}.311\,{\scriptstyle\pm.015}$ & $16.5\,{\scriptstyle\pm0.9}$ & $4.8\,{\scriptstyle\pm0.7}$ & $.314\,{\scriptstyle\pm.012}$ \\
\addlinespace[2pt]
\multicolumn{8}{@{}l}{\itshape CNN-GRU} \\
MAP & $86.4\,{\scriptstyle\pm1.4}$ & $77.1\,{\scriptstyle\pm3.2}$ & $73.8\,{\scriptstyle\pm2.1}$ & ${\scriptstyle-}.423\,{\scriptstyle\pm.064}$ & $21.5\,{\scriptstyle\pm2.2}$ & $7.1\,{\scriptstyle\pm0.7}$ & $.235\,{\scriptstyle\pm.054}$ \\
Temp. & $86.4\,{\scriptstyle\pm1.4}$ & $77.1\,{\scriptstyle\pm3.2}$ & $73.8\,{\scriptstyle\pm2.1}$ & ${\scriptstyle-}.396\,{\scriptstyle\pm.053}$ & $21.0\,{\scriptstyle\pm2.2}$ & $7.1\,{\scriptstyle\pm0.7}$ & $.311\,{\scriptstyle\pm.053}$ \\
SWA & $85.9\,{\scriptstyle\pm1.6}$ & $77.2\,{\scriptstyle\pm2.7}$ & $72.8\,{\scriptstyle\pm2.9}$ & ${\scriptstyle-}.396\,{\scriptstyle\pm.048}$ & $21.2\,{\scriptstyle\pm2.4}$ & $7.1\,{\scriptstyle\pm1.1}$ & $.320\,{\scriptstyle\pm.020}$ \\
SWAG & $86.8\,{\scriptstyle\pm1.4}$ & $78.8\,{\scriptstyle\pm2.2}$ & $74.6\,{\scriptstyle\pm2.6}$ & ${\scriptstyle-}.361\,{\scriptstyle\pm.037}$ & $19.5\,{\scriptstyle\pm2.1}$ & $7.0\,{\scriptstyle\pm1.2}$ & $.343\,{\scriptstyle\pm.026}$ \\
Laplace & $86.4\,{\scriptstyle\pm1.4}$ & $77.1\,{\scriptstyle\pm3.2}$ & $73.8\,{\scriptstyle\pm2.1}$ & ${\scriptstyle-}.423\,{\scriptstyle\pm.064}$ & $21.5\,{\scriptstyle\pm2.2}$ & $7.1\,{\scriptstyle\pm0.7}$ & $.235\,{\scriptstyle\pm.054}$ \\
VGD & $87.9\,{\scriptstyle\pm0.5}$ & $79.2\,{\scriptstyle\pm1.6}$ & $76.3\,{\scriptstyle\pm0.4}$ & ${\scriptstyle-}.356\,{\scriptstyle\pm.055}$ & $18.8\,{\scriptstyle\pm1.8}$ & $7.6\,{\scriptstyle\pm1.6}$ & $.348\,{\scriptstyle\pm.016}$ \\
MFVI & $87.4\,{\scriptstyle\pm0.4}$ & $77.4\,{\scriptstyle\pm3.4}$ & $75.1\,{\scriptstyle\pm0.8}$ & ${\scriptstyle-}.390\,{\scriptstyle\pm.016}$ & $20.2\,{\scriptstyle\pm0.4}$ & $8.2\,{\scriptstyle\pm1.2}$ & $.340\,{\scriptstyle\pm.045}$ \\
SGLD & $87.2\,{\scriptstyle\pm0.4}$ & $78.7\,{\scriptstyle\pm1.2}$ & $74.9\,{\scriptstyle\pm0.7}$ & ${\scriptstyle-}.375\,{\scriptstyle\pm.036}$ & $19.7\,{\scriptstyle\pm0.9}$ & $6.7\,{\scriptstyle\pm0.6}$ & $.263\,{\scriptstyle\pm.038}$ \\
DE & $\underline{88.0}\,{\scriptstyle\pm0.1}$ & $\underline{79.7}\,{\scriptstyle\pm0.7}$ & $\underline{76.3}\,{\scriptstyle\pm0.3}$ & ${\scriptstyle-}\underline{.347}\,{\scriptstyle\pm.018}$ & $\underline{18.7}\,{\scriptstyle\pm0.5}$ & $\underline{6.6}\,{\scriptstyle\pm0.4}$ & $.276\,{\scriptstyle\pm.023}$ \\
MultiSWA & $85.8\,{\scriptstyle\pm0.3}$ & $77.7\,{\scriptstyle\pm0.5}$ & $72.7\,{\scriptstyle\pm0.5}$ & ${\scriptstyle-}.382\,{\scriptstyle\pm.009}$ & $20.9\,{\scriptstyle\pm0.5}$ & $7.2\,{\scriptstyle\pm0.2}$ & $.325\,{\scriptstyle\pm.014}$ \\
MultiSWAG & $86.8\,{\scriptstyle\pm0.2}$ & $78.9\,{\scriptstyle\pm0.5}$ & $74.4\,{\scriptstyle\pm0.4}$ & ${\scriptstyle-}.352\,{\scriptstyle\pm.007}$ & $19.3\,{\scriptstyle\pm0.4}$ & $7.2\,{\scriptstyle\pm0.2}$ & $.343\,{\scriptstyle\pm.012}$ \\
\addlinespace[2pt]
\multicolumn{8}{@{}l}{\itshape Other baselines} \\
Somnotate & $80.9$ & $72.1$ & $68.7$ & ${\scriptstyle-}.863$ & $32.0$ & $14.4$ & $.110$ \\
Spindle & $\underline{88.6}\,{\scriptstyle\pm0.8}$ & $\underline{79.6}\,{\scriptstyle\pm1.2}$ & $\textbf{\underline{79.7}}\,{\scriptstyle\pm1.3}$ & ${\scriptstyle-}\underline{.676}\,{\scriptstyle\pm.034}$ & $\underline{19.9}\,{\scriptstyle\pm1.4}$ & $\underline{8.9}\,{\scriptstyle\pm0.6}$ & $.060\,{\scriptstyle\pm.005}$ \\
\bottomrule
\end{tabular}
\end{table*}                % \ref{tab:ood-results-wide}

%%% RANK
\begin{table*}[ht]
\centering
\caption{\textbf{Average MF1 rank.} Models were ranked jointly within each cohort, and ranks were averaged over cohorts and backbones ($1=$ best; lower is better). Other baselines are ranked individually. \textbf{Bold} and \underline{underlining} denote the best overall and within each model group, respectively.}
\label{tab:rank-averaged-transposed-mf1}
\scriptsize
\setlength{\tabcolsep}{3pt}
\begin{tabular*}{\textwidth}{@{\extracolsep{\fill}}lccccccccccccc@{}}
\toprule
 & \multicolumn{8}{c}{Single solution/trajectory} & \multicolumn{3}{c}{Multi-basin} & \multicolumn{2}{c}{Other baselines} \\
\cmidrule(lr){2-9} \cmidrule(lr){10-12} \cmidrule(lr){13-14}
Scope & MAP & Temp. & SWA & SWAG & Laplace & VGD & MFVI & SGLD & DE & MultiSWA & MultiSWAG & Somnotate & Spindle \\
\midrule
ID & 19.5 & 19.5 & 8.0 & 7.0 & 21.0 & 13.0 & 15.0 & 10.5 & 9.5 & \underline{3.5} & \textbf{1.5} & 20.0 & 24.0 \\
OOD & 16.6 & 16.6 & 12.9 & 10.1 & 17.2 & 13.6 & 13.6 & 10.9 & \underline{7.4} & 8.4 & \textbf{7.0} & 16.8 & 14.6 \\
ID + OOD & 17.1 & 17.1 & 12.1 & 9.6 & 17.8 & 13.5 & 13.8 & 10.8 & 7.8 & \underline{7.6} & \textbf{6.1} & 17.3 & 16.2 \\
\bottomrule
\end{tabular*}
\end{table*} 

%%% OOD DETECTION
\begin{table*}[t]
\centering
\caption{\textbf{Detection and rejection by total, epistemic, and aleatoric uncertainty.} Total uncertainty (TU) is the entropy of the predictive distribution, aleatoric uncertainty (AU) the mean entropy of the posterior samples or ensemble members it averages, and epistemic uncertainty (EU\,=\,TU\,$-$\,AU) their mutual information; the TU columns are those of Table~\ref{tab:uncertainty-utility}. AUROC measures detection of OOD recordings, artifact-containing epochs, and inter-rater disagreement; higher is better. OOD detection uses the per-recording mean of the epoch-level score to distinguish WT from the five held-out OOD cohorts, whereas artifact and disagreement detection use the epoch-level score on the \textsc{spindle} cohorts. AURC measures selective prediction across the OOD cohorts with the score as the rejection criterion; lower is better. Methods with a single predictive component (MAP, Temp., SWA, Somnotate, Spindle) have EU\,=\,0 and AU\,=\,TU: their EU columns are the chance-level AUROC (.50) and the AURC of rejecting in random order. Results are mean $\pm$ 95\% intervals across training seeds; for DE, MultiSWA, and MultiSWAG, intervals are the 2.5--97.5\% empirical interval across 100 resampled 10-member ensembles. \textbf{Bold} and \underline{underlining} denote the best overall and within each model group, respectively.}
\label{tab:uncertainty-decomposition}
\scriptsize
\setlength{\tabcolsep}{4pt}
\begin{tabular}{lccc|ccc|ccc|ccc}
\toprule
 & \multicolumn{9}{c}{AUROC $\uparrow$} & \multicolumn{3}{c}{AURC $\downarrow$} \\
\cmidrule(lr){2-10} \cmidrule(lr){11-13}
 & \multicolumn{3}{c}{OOD} & \multicolumn{3}{c}{Artifact} & \multicolumn{3}{c}{Disagree} & \multicolumn{3}{c}{Reject} \\
\cmidrule(lr){2-4} \cmidrule(lr){5-7} \cmidrule(lr){8-10} \cmidrule(lr){11-13}
Model & TU & EU & AU & TU & EU & AU & TU & EU & AU & TU & EU & AU \\
\midrule
\multicolumn{13}{@{}l}{\itshape U-Sleep} \\
MAP & $.72\,{\scriptstyle\pm.06}$ & $.50\,{\scriptstyle\pm.00}$ & $.72\,{\scriptstyle\pm.06}$ & $.44\,{\scriptstyle\pm.03}$ & $.50\,{\scriptstyle\pm.00}$ & $.44\,{\scriptstyle\pm.03}$ & $.76\,{\scriptstyle\pm.02}$ & $.50\,{\scriptstyle\pm.00}$ & $.76\,{\scriptstyle\pm.02}$ & $.054\,{\scriptstyle\pm.015}$ & $.152\,{\scriptstyle\pm.035}$ & $.054\,{\scriptstyle\pm.015}$ \\
Temp. & $.71\,{\scriptstyle\pm.06}$ & $.50\,{\scriptstyle\pm.00}$ & $.71\,{\scriptstyle\pm.06}$ & $.44\,{\scriptstyle\pm.03}$ & $.50\,{\scriptstyle\pm.00}$ & $.44\,{\scriptstyle\pm.03}$ & $.76\,{\scriptstyle\pm.02}$ & $.50\,{\scriptstyle\pm.00}$ & $.76\,{\scriptstyle\pm.02}$ & $.054\,{\scriptstyle\pm.015}$ & $.152\,{\scriptstyle\pm.035}$ & $.054\,{\scriptstyle\pm.015}$ \\
SWA & $.80\,{\scriptstyle\pm.03}$ & $.50\,{\scriptstyle\pm.00}$ & $.80\,{\scriptstyle\pm.03}$ & $.49\,{\scriptstyle\pm.03}$ & $.50\,{\scriptstyle\pm.00}$ & $.49\,{\scriptstyle\pm.03}$ & $.76\,{\scriptstyle\pm.01}$ & $.50\,{\scriptstyle\pm.00}$ & $.76\,{\scriptstyle\pm.01}$ & $.049\,{\scriptstyle\pm.019}$ & $.139\,{\scriptstyle\pm.031}$ & $.049\,{\scriptstyle\pm.019}$ \\
SWAG & $.82\,{\scriptstyle\pm.02}$ & $.79\,{\scriptstyle\pm.04}$ & $.80\,{\scriptstyle\pm.04}$ & $.51\,{\scriptstyle\pm.03}$ & $.58\,{\scriptstyle\pm.07}$ & $.50\,{\scriptstyle\pm.03}$ & $.75\,{\scriptstyle\pm.01}$ & $.67\,{\scriptstyle\pm.02}$ & $.77\,{\scriptstyle\pm.01}$ & $.041\,{\scriptstyle\pm.014}$ & $.042\,{\scriptstyle\pm.008}$ & $.043\,{\scriptstyle\pm.016}$ \\
Laplace & $.72\,{\scriptstyle\pm.06}$ & $.69\,{\scriptstyle\pm.05}$ & $.72\,{\scriptstyle\pm.06}$ & $.44\,{\scriptstyle\pm.03}$ & $.46\,{\scriptstyle\pm.02}$ & $.44\,{\scriptstyle\pm.03}$ & $.76\,{\scriptstyle\pm.02}$ & $\mathbf{\underline{.74}}\,{\scriptstyle\pm.02}$ & $.76\,{\scriptstyle\pm.02}$ & $.054\,{\scriptstyle\pm.015}$ & $.054\,{\scriptstyle\pm.015}$ & $.054\,{\scriptstyle\pm.015}$ \\
VGD & $.72\,{\scriptstyle\pm.07}$ & $.67\,{\scriptstyle\pm.09}$ & $.71\,{\scriptstyle\pm.07}$ & $.59\,{\scriptstyle\pm.08}$ & $.62\,{\scriptstyle\pm.09}$ & $.58\,{\scriptstyle\pm.07}$ & $.72\,{\scriptstyle\pm.05}$ & $.63\,{\scriptstyle\pm.06}$ & $.74\,{\scriptstyle\pm.04}$ & $.076\,{\scriptstyle\pm.069}$ & $.079\,{\scriptstyle\pm.059}$ & $.083\,{\scriptstyle\pm.081}$ \\
MFVI & $.71\,{\scriptstyle\pm.07}$ & $.65\,{\scriptstyle\pm.08}$ & $.73\,{\scriptstyle\pm.07}$ & $.65\,{\scriptstyle\pm.03}$ & $.67\,{\scriptstyle\pm.01}$ & $\underline{.64}\,{\scriptstyle\pm.04}$ & $.71\,{\scriptstyle\pm.01}$ & $.62\,{\scriptstyle\pm.03}$ & $.74\,{\scriptstyle\pm.02}$ & $.052\,{\scriptstyle\pm.013}$ & $.060\,{\scriptstyle\pm.014}$ & $.057\,{\scriptstyle\pm.017}$ \\
SGLD & $.78\,{\scriptstyle\pm.02}$ & $.77\,{\scriptstyle\pm.04}$ & $.77\,{\scriptstyle\pm.02}$ & $\underline{.66}\,{\scriptstyle\pm.02}$ & $\underline{.70}\,{\scriptstyle\pm.04}$ & $.64\,{\scriptstyle\pm.03}$ & $.73\,{\scriptstyle\pm.01}$ & $.63\,{\scriptstyle\pm.03}$ & $.76\,{\scriptstyle\pm.01}$ & $.041\,{\scriptstyle\pm.005}$ & $.048\,{\scriptstyle\pm.003}$ & $.042\,{\scriptstyle\pm.006}$ \\
DE & $.81\,{\scriptstyle\pm.03}$ & $.80\,{\scriptstyle\pm.04}$ & $.79\,{\scriptstyle\pm.03}$ & $.51\,{\scriptstyle\pm.05}$ & $.54\,{\scriptstyle\pm.07}$ & $.50\,{\scriptstyle\pm.04}$ & $\underline{.77}\,{\scriptstyle\pm.02}$ & $.69\,{\scriptstyle\pm.03}$ & $\mathbf{\underline{.80}}\,{\scriptstyle\pm.01}$ & $.027\,{\scriptstyle\pm.005}$ & $\mathbf{\underline{.031}}\,{\scriptstyle\pm.005}$ & $.033\,{\scriptstyle\pm.006}$ \\
MultiSWA & $.83\,{\scriptstyle\pm.01}$ & $\underline{.85}\,{\scriptstyle\pm.02}$ & $.81\,{\scriptstyle\pm.01}$ & $.53\,{\scriptstyle\pm.03}$ & $.58\,{\scriptstyle\pm.04}$ & $.51\,{\scriptstyle\pm.03}$ & $.76\,{\scriptstyle\pm.01}$ & $.67\,{\scriptstyle\pm.02}$ & $.78\,{\scriptstyle\pm.01}$ & $.028\,{\scriptstyle\pm.004}$ & $.034\,{\scriptstyle\pm.004}$ & $.033\,{\scriptstyle\pm.006}$ \\
MultiSWAG & $\underline{.83}\,{\scriptstyle\pm.01}$ & $.81\,{\scriptstyle\pm.02}$ & $\underline{.82}\,{\scriptstyle\pm.01}$ & $.56\,{\scriptstyle\pm.03}$ & $.63\,{\scriptstyle\pm.04}$ & $.53\,{\scriptstyle\pm.03}$ & $.76\,{\scriptstyle\pm.01}$ & $.66\,{\scriptstyle\pm.02}$ & $.79\,{\scriptstyle\pm.01}$ & $\mathbf{\underline{.026}}\,{\scriptstyle\pm.004}$ & $.032\,{\scriptstyle\pm.004}$ & $\mathbf{\underline{.032}}\,{\scriptstyle\pm.006}$ \\
\addlinespace[2pt]
\multicolumn{13}{@{}l}{\itshape CNN-GRU} \\
MAP & $.75\,{\scriptstyle\pm.04}$ & $.50\,{\scriptstyle\pm.00}$ & $.75\,{\scriptstyle\pm.04}$ & $.54\,{\scriptstyle\pm.04}$ & $.50\,{\scriptstyle\pm.00}$ & $.54\,{\scriptstyle\pm.04}$ & $.75\,{\scriptstyle\pm.02}$ & $.50\,{\scriptstyle\pm.00}$ & $.75\,{\scriptstyle\pm.02}$ & $.055\,{\scriptstyle\pm.017}$ & $.134\,{\scriptstyle\pm.014}$ & $.055\,{\scriptstyle\pm.017}$ \\
Temp. & $.77\,{\scriptstyle\pm.04}$ & $.50\,{\scriptstyle\pm.00}$ & $.77\,{\scriptstyle\pm.04}$ & $.54\,{\scriptstyle\pm.04}$ & $.50\,{\scriptstyle\pm.00}$ & $.54\,{\scriptstyle\pm.04}$ & $.75\,{\scriptstyle\pm.02}$ & $.50\,{\scriptstyle\pm.00}$ & $.75\,{\scriptstyle\pm.02}$ & $.055\,{\scriptstyle\pm.017}$ & $.134\,{\scriptstyle\pm.014}$ & $.055\,{\scriptstyle\pm.017}$ \\
SWA & $.84\,{\scriptstyle\pm.03}$ & $.50\,{\scriptstyle\pm.00}$ & $.84\,{\scriptstyle\pm.03}$ & $.70\,{\scriptstyle\pm.03}$ & $.50\,{\scriptstyle\pm.00}$ & $.70\,{\scriptstyle\pm.03}$ & $.72\,{\scriptstyle\pm.02}$ & $.50\,{\scriptstyle\pm.00}$ & $.72\,{\scriptstyle\pm.02}$ & $.064\,{\scriptstyle\pm.025}$ & $.138\,{\scriptstyle\pm.016}$ & $.064\,{\scriptstyle\pm.025}$ \\
SWAG & $.83\,{\scriptstyle\pm.02}$ & $.79\,{\scriptstyle\pm.03}$ & $.83\,{\scriptstyle\pm.02}$ & $.69\,{\scriptstyle\pm.03}$ & $\mathbf{\underline{.74}}\,{\scriptstyle\pm.02}$ & $.68\,{\scriptstyle\pm.03}$ & $.74\,{\scriptstyle\pm.02}$ & $.60\,{\scriptstyle\pm.03}$ & $.75\,{\scriptstyle\pm.02}$ & $.052\,{\scriptstyle\pm.015}$ & $.056\,{\scriptstyle\pm.010}$ & $.056\,{\scriptstyle\pm.020}$ \\
Laplace & $.75\,{\scriptstyle\pm.04}$ & $.79\,{\scriptstyle\pm.04}$ & $.75\,{\scriptstyle\pm.04}$ & $.54\,{\scriptstyle\pm.04}$ & $.60\,{\scriptstyle\pm.04}$ & $.54\,{\scriptstyle\pm.04}$ & $.75\,{\scriptstyle\pm.02}$ & $.68\,{\scriptstyle\pm.04}$ & $.75\,{\scriptstyle\pm.02}$ & $.055\,{\scriptstyle\pm.017}$ & $.053\,{\scriptstyle\pm.013}$ & $.055\,{\scriptstyle\pm.017}$ \\
VGD & $.81\,{\scriptstyle\pm.02}$ & $.71\,{\scriptstyle\pm.06}$ & $.82\,{\scriptstyle\pm.02}$ & $.59\,{\scriptstyle\pm.06}$ & $.57\,{\scriptstyle\pm.09}$ & $.60\,{\scriptstyle\pm.06}$ & $.77\,{\scriptstyle\pm.01}$ & $.70\,{\scriptstyle\pm.03}$ & $.77\,{\scriptstyle\pm.01}$ & $.049\,{\scriptstyle\pm.026}$ & $.047\,{\scriptstyle\pm.021}$ & $.051\,{\scriptstyle\pm.027}$ \\
MFVI & $.73\,{\scriptstyle\pm.05}$ & $.69\,{\scriptstyle\pm.06}$ & $.74\,{\scriptstyle\pm.04}$ & $.50\,{\scriptstyle\pm.05}$ & $.50\,{\scriptstyle\pm.03}$ & $.51\,{\scriptstyle\pm.05}$ & $\mathbf{\underline{.77}}\,{\scriptstyle\pm.01}$ & $.71\,{\scriptstyle\pm.02}$ & $\underline{.78}\,{\scriptstyle\pm.01}$ & $.050\,{\scriptstyle\pm.009}$ & $.047\,{\scriptstyle\pm.009}$ & $.056\,{\scriptstyle\pm.009}$ \\
SGLD & $.76\,{\scriptstyle\pm.05}$ & $.79\,{\scriptstyle\pm.05}$ & $.74\,{\scriptstyle\pm.05}$ & $.54\,{\scriptstyle\pm.04}$ & $.55\,{\scriptstyle\pm.04}$ & $.54\,{\scriptstyle\pm.04}$ & $.76\,{\scriptstyle\pm.01}$ & $\underline{.71}\,{\scriptstyle\pm.02}$ & $.77\,{\scriptstyle\pm.01}$ & $.042\,{\scriptstyle\pm.013}$ & $\underline{.040}\,{\scriptstyle\pm.010}$ & $.044\,{\scriptstyle\pm.013}$ \\
DE & $.81\,{\scriptstyle\pm.02}$ & $.83\,{\scriptstyle\pm.03}$ & $.79\,{\scriptstyle\pm.02}$ & $.57\,{\scriptstyle\pm.03}$ & $.57\,{\scriptstyle\pm.04}$ & $.57\,{\scriptstyle\pm.03}$ & $.76\,{\scriptstyle\pm.01}$ & $.69\,{\scriptstyle\pm.02}$ & $.77\,{\scriptstyle\pm.01}$ & $\underline{.036}\,{\scriptstyle\pm.006}$ & $.041\,{\scriptstyle\pm.009}$ & $\underline{.037}\,{\scriptstyle\pm.006}$ \\
MultiSWA & $\underline{.85}\,{\scriptstyle\pm.01}$ & $\mathbf{\underline{.85}}\,{\scriptstyle\pm.02}$ & $\underline{.84}\,{\scriptstyle\pm.01}$ & $\mathbf{\underline{.71}}\,{\scriptstyle\pm.01}$ & $.73\,{\scriptstyle\pm.02}$ & $\mathbf{\underline{.70}}\,{\scriptstyle\pm.01}$ & $.73\,{\scriptstyle\pm.01}$ & $.59\,{\scriptstyle\pm.04}$ & $.73\,{\scriptstyle\pm.01}$ & $.055\,{\scriptstyle\pm.006}$ & $.056\,{\scriptstyle\pm.007}$ & $.057\,{\scriptstyle\pm.006}$ \\
MultiSWAG & $.84\,{\scriptstyle\pm.01}$ & $.80\,{\scriptstyle\pm.02}$ & $.83\,{\scriptstyle\pm.01}$ & $.69\,{\scriptstyle\pm.01}$ & $.74\,{\scriptstyle\pm.01}$ & $.68\,{\scriptstyle\pm.01}$ & $.74\,{\scriptstyle\pm.01}$ & $.61\,{\scriptstyle\pm.02}$ & $.76\,{\scriptstyle\pm.01}$ & $.048\,{\scriptstyle\pm.004}$ & $.048\,{\scriptstyle\pm.003}$ & $.052\,{\scriptstyle\pm.005}$ \\
\addlinespace[2pt]
\multicolumn{13}{@{}l}{\itshape Other baselines} \\
Somnotate & $.86$ & $.50$ & $.86$ & $\underline{.44}$ & $.50$ & $\underline{.44}$ & $.71$ & $.50$ & $.71$ & $.072$ & $.188$ & $.072$ \\
Spindle & $\mathbf{\underline{.88}}\,{\scriptstyle\pm.01}$ & $.50\,{\scriptstyle\pm.00}$ & $\mathbf{\underline{.88}}\,{\scriptstyle\pm.01}$ & $.40\,{\scriptstyle\pm.04}$ & $.50\,{\scriptstyle\pm.00}$ & $.40\,{\scriptstyle\pm.04}$ & $\underline{.76}\,{\scriptstyle\pm.01}$ & $.50\,{\scriptstyle\pm.00}$ & $\underline{.76}\,{\scriptstyle\pm.01}$ & $\underline{.043}\,{\scriptstyle\pm.003}$ & $\underline{.111}\,{\scriptstyle\pm.008}$ & $\underline{.043}\,{\scriptstyle\pm.003}$ \\
\bottomrule
\end{tabular}
\end{table*}
\begin{table*}[t]
\centering
\caption{\textbf{Predictive entropy and uncertainty utility.} Mean predictive entropy $H$ (nats) is shown for WT (ID), the mean over the five OOD cohorts, and \textsc{spindle} windows grouped by scorer agreement and artifact flags. Agree/Disagree excludes artifact-flagged windows; Artifact denotes a flag by either scorer. AUROC evaluates OOD-recording detection using per-recording mean entropy and disagreement/artifact detection using window-level entropy. AURC evaluates entropy-based selective prediction on the OOD cohorts; lower is better. Values are mean $\pm$ 95\% CIs across seeds; for DE, MultiSWA, and MultiSWAG, $\pm$ denotes half the empirical 2.5--97.5\% interval across 100 resampled 10-member ensembles. \textbf{Bold} and \underline{underlining} denote the best overall and within each model group, respectively.}
\label{tab:uncertainty-utility-full}
\scriptsize
\setlength{\tabcolsep}{4pt}
\begin{tabular}{lcc|cc|cccccc}
\toprule
 & \multicolumn{6}{c}{Entropy ($H$)} & \multicolumn{3}{c}{Detection AUROC $\uparrow$} & \multicolumn{1}{c}{AURC $\downarrow$} \\
\cmidrule(lr){2-7} \cmidrule(lr){8-10} \cmidrule(lr){11-11}
Model & ID & OOD & Agree & Disagree & No artifact & Artifact & OOD & Disagree & Artifact & OOD \\
\midrule
\multicolumn{11}{@{}l}{\itshape U-Sleep} \\
MAP & $.17\,{\scriptstyle\pm.04}$ & $.22\,{\scriptstyle\pm.04}$ & $.22\,{\scriptstyle\pm.04}$ & $.48\,{\scriptstyle\pm.08}$ & $.24\,{\scriptstyle\pm.04}$ & $.20\,{\scriptstyle\pm.03}$ & $.72\,{\scriptstyle\pm.06}$ & $.76\,{\scriptstyle\pm.02}$ & $.44\,{\scriptstyle\pm.03}$ & $.054\,{\scriptstyle\pm.015}$ \\
Temp. & $.19\,{\scriptstyle\pm.02}$ & $.25\,{\scriptstyle\pm.02}$ & $.24\,{\scriptstyle\pm.02}$ & $.51\,{\scriptstyle\pm.05}$ & $.26\,{\scriptstyle\pm.02}$ & $.22\,{\scriptstyle\pm.01}$ & $.71\,{\scriptstyle\pm.06}$ & $.76\,{\scriptstyle\pm.02}$ & $.44\,{\scriptstyle\pm.03}$ & $.054\,{\scriptstyle\pm.015}$ \\
SWA & $.15\,{\scriptstyle\pm.01}$ & $.24\,{\scriptstyle\pm.03}$ & $.24\,{\scriptstyle\pm.03}$ & $.51\,{\scriptstyle\pm.05}$ & $.26\,{\scriptstyle\pm.03}$ & $.26\,{\scriptstyle\pm.02}$ & $.80\,{\scriptstyle\pm.03}$ & $.76\,{\scriptstyle\pm.01}$ & $.49\,{\scriptstyle\pm.03}$ & $.049\,{\scriptstyle\pm.019}$ \\
SWAG & $.16\,{\scriptstyle\pm.01}$ & $.28\,{\scriptstyle\pm.03}$ & $.28\,{\scriptstyle\pm.03}$ & $.56\,{\scriptstyle\pm.05}$ & $.30\,{\scriptstyle\pm.03}$ & $.31\,{\scriptstyle\pm.01}$ & $.82\,{\scriptstyle\pm.02}$ & $.75\,{\scriptstyle\pm.01}$ & $.51\,{\scriptstyle\pm.03}$ & $.041\,{\scriptstyle\pm.014}$ \\
Laplace & $.17\,{\scriptstyle\pm.04}$ & $.22\,{\scriptstyle\pm.04}$ & $.22\,{\scriptstyle\pm.04}$ & $.48\,{\scriptstyle\pm.08}$ & $.24\,{\scriptstyle\pm.04}$ & $.20\,{\scriptstyle\pm.03}$ & $.72\,{\scriptstyle\pm.06}$ & $.76\,{\scriptstyle\pm.02}$ & $.44\,{\scriptstyle\pm.03}$ & $.054\,{\scriptstyle\pm.015}$ \\
VGD & $.25\,{\scriptstyle\pm.04}$ & $.39\,{\scriptstyle\pm.04}$ & $.37\,{\scriptstyle\pm.04}$ & $.64\,{\scriptstyle\pm.06}$ & $.39\,{\scriptstyle\pm.03}$ & $.48\,{\scriptstyle\pm.11}$ & $.72\,{\scriptstyle\pm.07}$ & $.72\,{\scriptstyle\pm.05}$ & $.59\,{\scriptstyle\pm.08}$ & $.076\,{\scriptstyle\pm.069}$ \\
MFVI & $.37\,{\scriptstyle\pm.04}$ & $.50\,{\scriptstyle\pm.05}$ & $.45\,{\scriptstyle\pm.06}$ & $.72\,{\scriptstyle\pm.05}$ & $.47\,{\scriptstyle\pm.06}$ & $.63\,{\scriptstyle\pm.07}$ & $.71\,{\scriptstyle\pm.07}$ & $.71\,{\scriptstyle\pm.01}$ & $.65\,{\scriptstyle\pm.03}$ & $.052\,{\scriptstyle\pm.013}$ \\
SGLD & $.23\,{\scriptstyle\pm.01}$ & $.36\,{\scriptstyle\pm.01}$ & $.33\,{\scriptstyle\pm.01}$ & $.61\,{\scriptstyle\pm.01}$ & $.35\,{\scriptstyle\pm.01}$ & $.50\,{\scriptstyle\pm.01}$ & $.78\,{\scriptstyle\pm.02}$ & $.73\,{\scriptstyle\pm.01}$ & $\underline{.66}\,{\scriptstyle\pm.02}$ & $.041\,{\scriptstyle\pm.005}$ \\
DE & $.18\,{\scriptstyle\pm.01}$ & $.32\,{\scriptstyle\pm.03}$ & $.32\,{\scriptstyle\pm.03}$ & $.63\,{\scriptstyle\pm.02}$ & $.34\,{\scriptstyle\pm.03}$ & $.34\,{\scriptstyle\pm.06}$ & $.81\,{\scriptstyle\pm.03}$ & $\underline{.77}\,{\scriptstyle\pm.02}$ & $.51\,{\scriptstyle\pm.05}$ & $.027\,{\scriptstyle\pm.005}$ \\
MultiSWA & $.15\,{\scriptstyle\pm.00}$ & $.29\,{\scriptstyle\pm.01}$ & $.29\,{\scriptstyle\pm.01}$ & $.59\,{\scriptstyle\pm.02}$ & $.31\,{\scriptstyle\pm.01}$ & $.33\,{\scriptstyle\pm.03}$ & $.83\,{\scriptstyle\pm.01}$ & $.76\,{\scriptstyle\pm.01}$ & $.53\,{\scriptstyle\pm.03}$ & $.028\,{\scriptstyle\pm.004}$ \\
MultiSWAG & $.16\,{\scriptstyle\pm.00}$ & $.31\,{\scriptstyle\pm.01}$ & $.31\,{\scriptstyle\pm.01}$ & $.61\,{\scriptstyle\pm.02}$ & $.33\,{\scriptstyle\pm.01}$ & $.38\,{\scriptstyle\pm.03}$ & $\underline{.83}\,{\scriptstyle\pm.01}$ & $.76\,{\scriptstyle\pm.01}$ & $.56\,{\scriptstyle\pm.03}$ & $\mathbf{\underline{.026}}\,{\scriptstyle\pm.004}$ \\
\addlinespace[2pt]
\multicolumn{11}{@{}l}{\itshape CNN-GRU} \\
MAP & $.12\,{\scriptstyle\pm.01}$ & $.23\,{\scriptstyle\pm.05}$ & $.24\,{\scriptstyle\pm.06}$ & $.50\,{\scriptstyle\pm.07}$ & $.26\,{\scriptstyle\pm.06}$ & $.26\,{\scriptstyle\pm.08}$ & $.75\,{\scriptstyle\pm.04}$ & $.75\,{\scriptstyle\pm.02}$ & $.54\,{\scriptstyle\pm.04}$ & $.055\,{\scriptstyle\pm.017}$ \\
Temp. & $.17\,{\scriptstyle\pm.01}$ & $.31\,{\scriptstyle\pm.05}$ & $.32\,{\scriptstyle\pm.06}$ & $.58\,{\scriptstyle\pm.06}$ & $.33\,{\scriptstyle\pm.06}$ & $.35\,{\scriptstyle\pm.08}$ & $.77\,{\scriptstyle\pm.04}$ & $.75\,{\scriptstyle\pm.02}$ & $.54\,{\scriptstyle\pm.04}$ & $.055\,{\scriptstyle\pm.017}$ \\
SWA & $.15\,{\scriptstyle\pm.02}$ & $.32\,{\scriptstyle\pm.02}$ & $.30\,{\scriptstyle\pm.03}$ & $.59\,{\scriptstyle\pm.06}$ & $.32\,{\scriptstyle\pm.03}$ & $.50\,{\scriptstyle\pm.02}$ & $.84\,{\scriptstyle\pm.03}$ & $.72\,{\scriptstyle\pm.02}$ & $.70\,{\scriptstyle\pm.03}$ & $.064\,{\scriptstyle\pm.025}$ \\
SWAG & $.17\,{\scriptstyle\pm.02}$ & $.34\,{\scriptstyle\pm.03}$ & $.33\,{\scriptstyle\pm.03}$ & $.63\,{\scriptstyle\pm.05}$ & $.35\,{\scriptstyle\pm.03}$ & $.52\,{\scriptstyle\pm.03}$ & $.83\,{\scriptstyle\pm.02}$ & $.74\,{\scriptstyle\pm.02}$ & $.69\,{\scriptstyle\pm.03}$ & $.052\,{\scriptstyle\pm.015}$ \\
Laplace & $.12\,{\scriptstyle\pm.01}$ & $.23\,{\scriptstyle\pm.05}$ & $.24\,{\scriptstyle\pm.06}$ & $.50\,{\scriptstyle\pm.07}$ & $.26\,{\scriptstyle\pm.06}$ & $.26\,{\scriptstyle\pm.08}$ & $.75\,{\scriptstyle\pm.04}$ & $.75\,{\scriptstyle\pm.02}$ & $.54\,{\scriptstyle\pm.04}$ & $.055\,{\scriptstyle\pm.017}$ \\
VGD & $.20\,{\scriptstyle\pm.01}$ & $.35\,{\scriptstyle\pm.02}$ & $.34\,{\scriptstyle\pm.03}$ & $.67\,{\scriptstyle\pm.04}$ & $.36\,{\scriptstyle\pm.03}$ & $.43\,{\scriptstyle\pm.04}$ & $.81\,{\scriptstyle\pm.02}$ & $.77\,{\scriptstyle\pm.01}$ & $.59\,{\scriptstyle\pm.06}$ & $.049\,{\scriptstyle\pm.026}$ \\
MFVI & $.24\,{\scriptstyle\pm.02}$ & $.34\,{\scriptstyle\pm.05}$ & $.35\,{\scriptstyle\pm.05}$ & $.67\,{\scriptstyle\pm.06}$ & $.37\,{\scriptstyle\pm.05}$ & $.35\,{\scriptstyle\pm.07}$ & $.73\,{\scriptstyle\pm.05}$ & $\mathbf{\underline{.77}}\,{\scriptstyle\pm.01}$ & $.50\,{\scriptstyle\pm.05}$ & $.050\,{\scriptstyle\pm.009}$ \\
SGLD & $.15\,{\scriptstyle\pm.01}$ & $.26\,{\scriptstyle\pm.04}$ & $.26\,{\scriptstyle\pm.04}$ & $.56\,{\scriptstyle\pm.04}$ & $.28\,{\scriptstyle\pm.04}$ & $.29\,{\scriptstyle\pm.06}$ & $.76\,{\scriptstyle\pm.05}$ & $.76\,{\scriptstyle\pm.01}$ & $.54\,{\scriptstyle\pm.04}$ & $.042\,{\scriptstyle\pm.013}$ \\
DE & $.14\,{\scriptstyle\pm.01}$ & $.28\,{\scriptstyle\pm.02}$ & $.27\,{\scriptstyle\pm.02}$ & $.56\,{\scriptstyle\pm.02}$ & $.29\,{\scriptstyle\pm.02}$ & $.33\,{\scriptstyle\pm.04}$ & $.81\,{\scriptstyle\pm.02}$ & $.76\,{\scriptstyle\pm.01}$ & $.57\,{\scriptstyle\pm.03}$ & $\underline{.036}\,{\scriptstyle\pm.006}$ \\
MultiSWA & $.15\,{\scriptstyle\pm.00}$ & $.33\,{\scriptstyle\pm.01}$ & $.31\,{\scriptstyle\pm.01}$ & $.59\,{\scriptstyle\pm.01}$ & $.33\,{\scriptstyle\pm.01}$ & $.51\,{\scriptstyle\pm.02}$ & $\underline{.85}\,{\scriptstyle\pm.01}$ & $.73\,{\scriptstyle\pm.01}$ & $\mathbf{\underline{.71}}\,{\scriptstyle\pm.01}$ & $.055\,{\scriptstyle\pm.006}$ \\
MultiSWAG & $.17\,{\scriptstyle\pm.00}$ & $.34\,{\scriptstyle\pm.01}$ & $.33\,{\scriptstyle\pm.01}$ & $.63\,{\scriptstyle\pm.01}$ & $.35\,{\scriptstyle\pm.01}$ & $.52\,{\scriptstyle\pm.02}$ & $.84\,{\scriptstyle\pm.01}$ & $.74\,{\scriptstyle\pm.01}$ & $.69\,{\scriptstyle\pm.01}$ & $.048\,{\scriptstyle\pm.004}$ \\
\addlinespace[2pt]
\multicolumn{11}{@{}l}{\itshape Other baselines} \\
Somnotate & $.03$ & $.11$ & $.13$ & $.25$ & $.14$ & $.08$ & $.86$ & $.71$ & $\underline{.44}$ & $.072$ \\
Spindle & $.04\,{\scriptstyle\pm.00}$ & $.06\,{\scriptstyle\pm.01}$ & $.06\,{\scriptstyle\pm.01}$ & $.20\,{\scriptstyle\pm.01}$ & $.07\,{\scriptstyle\pm.01}$ & $.04\,{\scriptstyle\pm.00}$ & $\mathbf{\underline{.88}}\,{\scriptstyle\pm.01}$ & $\underline{.76}\,{\scriptstyle\pm.01}$ & $.40\,{\scriptstyle\pm.04}$ & $\underline{.043}\,{\scriptstyle\pm.003}$ \\
\bottomrule
\end{tabular}
\end{table*}     % \ref{tab:ood-detection}

\begin{figure}[ht]
    \centering
    \includegraphics[width=\linewidth]{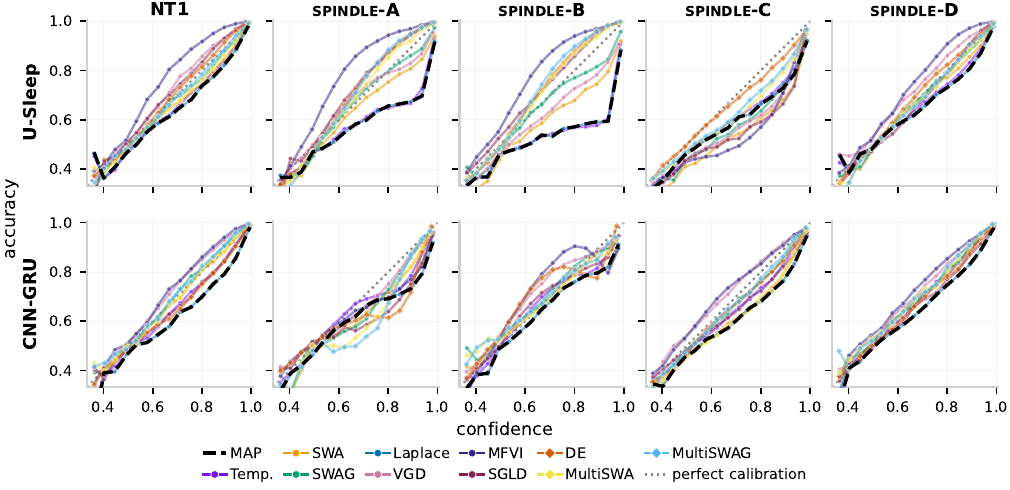}
    \caption{\textbf{Reliability curves per OOD cohort.} Columns are the five held-out OOD cohorts (NT1, \textsc{spindle}-A--D) and rows the two backbones. Each point is a confidence bin, plotted at its mean confidence against the accuracy of its epochs; the dotted diagonal is perfect calibration, and points below it indicate overconfidence. Curves are averaged across training seeds; for DE, MultiSWA, and MultiSWAG, across 100 resampled 10-member ensembles. MAP is shown as the dashed black reference.}
    \label{fig:reliability-ood-cohorts}
\end{figure}

\begin{figure}[ht]
    \centering
    \includegraphics[width=\linewidth]{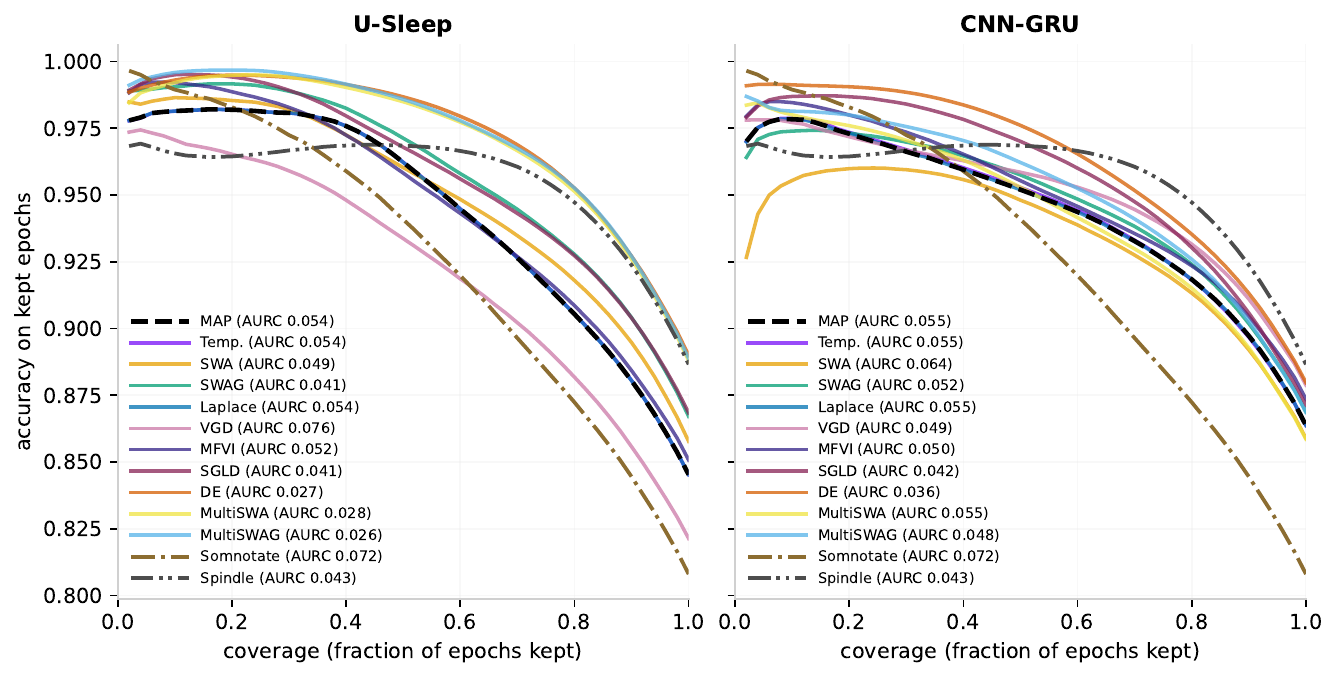}
    \caption{\textbf{Selective prediction on the OOD cohorts.} Within each held-out OOD cohort, epochs are rejected in order of decreasing predictive entropy; curves show the accuracy of the retained epochs against the retained fraction (coverage), averaged over the five cohorts. The legend reports the AURC, the area under the corresponding risk--coverage curve with risk $=1-$accuracy, as in Table~\ref{tab:uncertainty-utility}; lower is better. Curves and AURC are averaged across training seeds; for DE, MultiSWA, and MultiSWAG, across 100 resampled 10-member ensembles. MAP is shown as the dashed black reference. Somnotate and Spindle do not depend on the backbone and appear in both panels (dash-dotted).}

    \label{fig:rejection-curves}
\end{figure}

\begin{figure}[ht]
    \centering
    \includegraphics[width=0.85\linewidth]{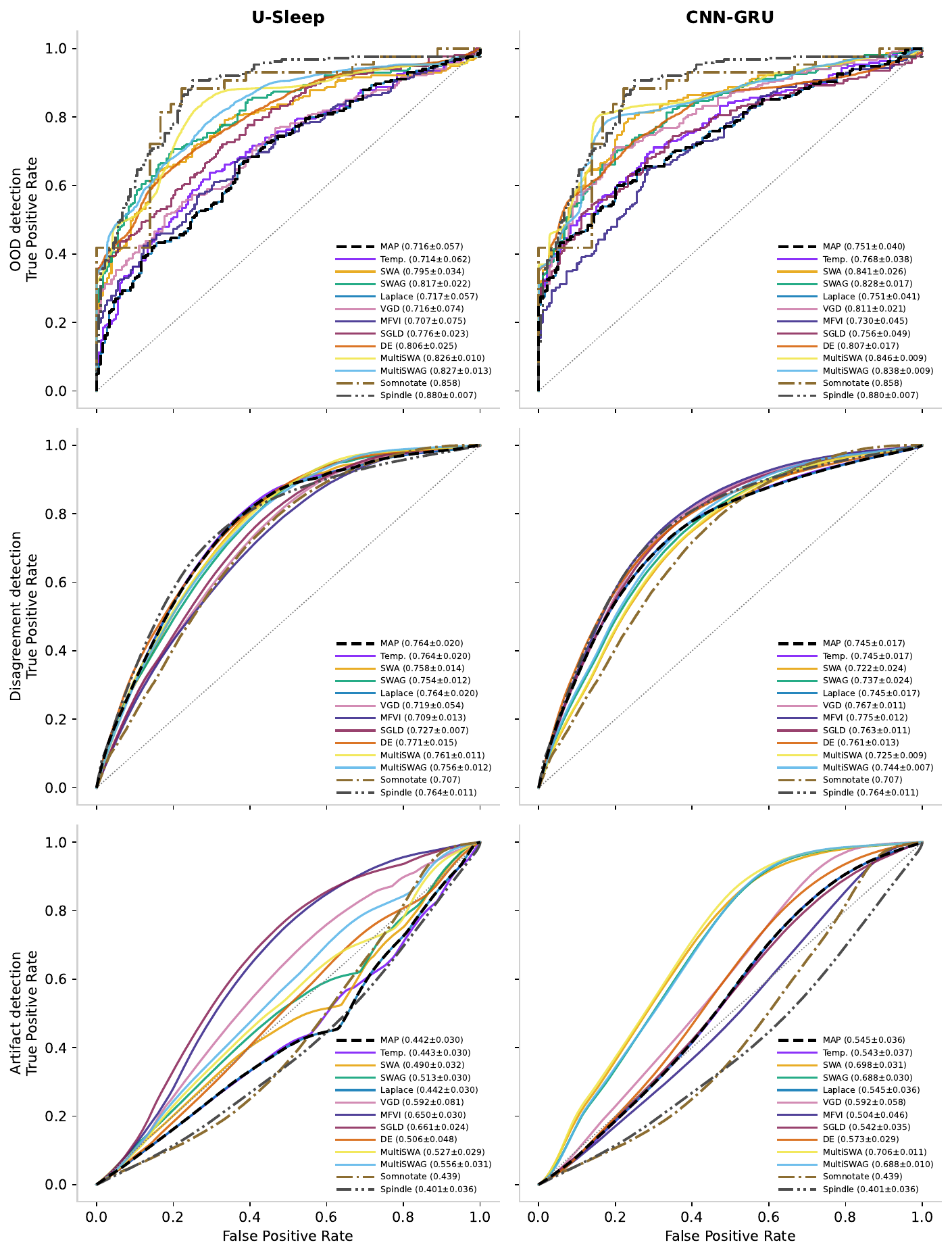}
    \caption{\textbf{Detection by predictive entropy.} ROC curves for OOD, disagreement, and artifact detection (rows) with predictive entropy as the score, for the two backbones (columns). OOD detection uses the per-recording mean entropy to distinguish WT test recordings from those of the five held-out OOD cohorts; disagreement and artifact detection use the epoch-level entropy on the \textsc{spindle} cohorts to identify epochs with inter-rater disagreement and artifact-containing epochs, respectively. Curves pool all training seeds (for DE, MultiSWA, and MultiSWAG, the 100 resampled 10-member ensembles); the legend reports the AUROC as in Table~\ref{tab:uncertainty-utility}, mean $\pm$ 95\% CI across seeds, where for the ensembles $\pm$ denotes half the empirical 2.5--97.5\% interval. MAP is shown as the dashed black reference, and the dotted diagonal is chance. Somnotate and Spindle do not depend on the backbone and appear in both columns (dash-dotted); Somnotate is a single model and has no interval.}
    \label{fig:ood-detection}
\end{figure}

\end{document}